\documentclass[runningheads]{llncs}

\usepackage{eccv}

\usepackage{eccvabbrv}
\usepackage{graphicx}
\usepackage{booktabs}
\usepackage{multirow}
\usepackage[accsupp]{axessibility}  
\usepackage[table,xcdraw]{xcolor}
\usepackage{colortbl}
\usepackage{pgfplots}
\usepackage[normalem]{ulem}
\usepackage{wrapfig}

\usepackage{hyperref}

\usepackage{orcidlink}

\begin{document}
\useunder{\uline}{\ul}{}

\title{EA-Swin: An Embedding-Agnostic Swin Transformer for AI-Generated Video Detection} 

\titlerunning{Embedding-Agnostic Swin Transformer for AI-Generated Video Detection}

\author{ Martin (Hung) Mai \inst{1,2} \orcidlink{0009-0006-9558-7052} \and
Loi Dinh \inst{3}  \and
Duc Hai Nguyen \inst{1,4} \orcidlink{0009-0007-6761-7020} \and
Dat Do \inst{5}\and
Luong Doan \inst{1} \orcidlink{0009-0008-5025-1883} \and
Khanh Nguyen Quoc \inst{1,4} \orcidlink{0009-0002-4453-1674} \and
Huan Vu \inst{2} \orcidlink{0000-0002-0785-6907} \and
Naeem Ul Islam $^{\ast}$  \inst{6} \orcidlink{0000-0002-7806-9601}  \and 
Tuan Do \thanks{Corresponding authors} \inst{1} \orcidlink{0009-0008-2111-0870}  }

\authorrunning{Mai et al.}

\institute{N2TP Technology Solution JSC, Hanoi, Vietnam 
\email{\{pqhung.mai,duchai.nguyen,luong.doan,quockhanh.nguyen,tuan.do\}@n2tp.com} \\ \and
College of Technology, National Economics University, Hanoi, Vietnam \and
University of Science, Vietnam National University, Ho Chi Minh City, Vietnam \and
Faculty of AI and Data Science, Phenikaa University, Hanoi, Vietnam \and
The Saigon International University, Ho Chi Minh City, Vietnam \and
College of Informatics, Yuan Ze University, Taoyuan, Taiwan
\email{naeem@saturn.yzu.edu.tw}
}

\maketitle

\begin{abstract}
Recent advances in foundation video generators such as Sora2, Veo3, and other commercial systems have produced highly realistic synthetic videos, exposing the limitations of existing detection methods that rely on shallow embedding trajectories, image-based adaptation, or computationally heavy MLLMs. We propose EA-Swin, an Embedding-Agnostic Swin Transformer that models spatiotemporal dependencies directly on pretrained video embeddings via a factorized windowed attention design, making it compatible with generic ViT-style patch-based encoders. Moreover, we construct the EA-Video dataset, a benchmark dataset comprising 130K videos that integrates newly collected samples with curated existing datasets, covering diverse commercial and open-source generators and including unseen-generator splits for rigorous cross-distribution evaluation. Extensive experiments show that EA-Swin achieves 0.97–0.99 accuracy across major generators, outperforming prior SoTA methods (typically 0.8–0.9) by a margin of 5–20\%, while maintaining strong generalization to unseen distributions, establishing a scalable and robust solution for modern AI-generated video detection.
\end{abstract}

\section{Introduction}
\label{sec:intro}
Recent advances in generative artificial intelligence have led to a rapid transformation in video synthesis capabilities. Early video generation models in 2023 \cite{chen2024videocrafter2,Khachatryan_2023_ICCV,wang2023modelscopetexttovideotechnicalreport} could only generate short, low-fidelity videos with limited temporal coherence. However, by 2025, hyper-realistic foundation models (e.g., Sora-2 \cite{sora2} by OpenAI and Veo-3 \cite{veo3} by Google) are capable of generating long, photorealistic videos from minimal input, including text prompts, reference images, or short video segments (see Figure \ref{fig:video_sample}). Powered by large-scale diffusion models \cite{rombach2022high}, transformers \cite{vaswani2017attention, dosovitskiy2020image}, and flow-matching \cite{lipman2022flow} techniques, these systems can synthesize content that is increasingly difficult to distinguish from real-world footage, even defeating human perceptual detection capabilities in some cases. This has raised significant concerns about the use of Generative AI with malicious intentions, such as generation of inappropriate content and large-scale visual media fabrication \cite{gambin2024deepfakes, YOON2025101491,easttom,CHEN2025108448}. 

\begin{figure}[!h]
  \centering
  \includegraphics[width=0.8\textwidth]{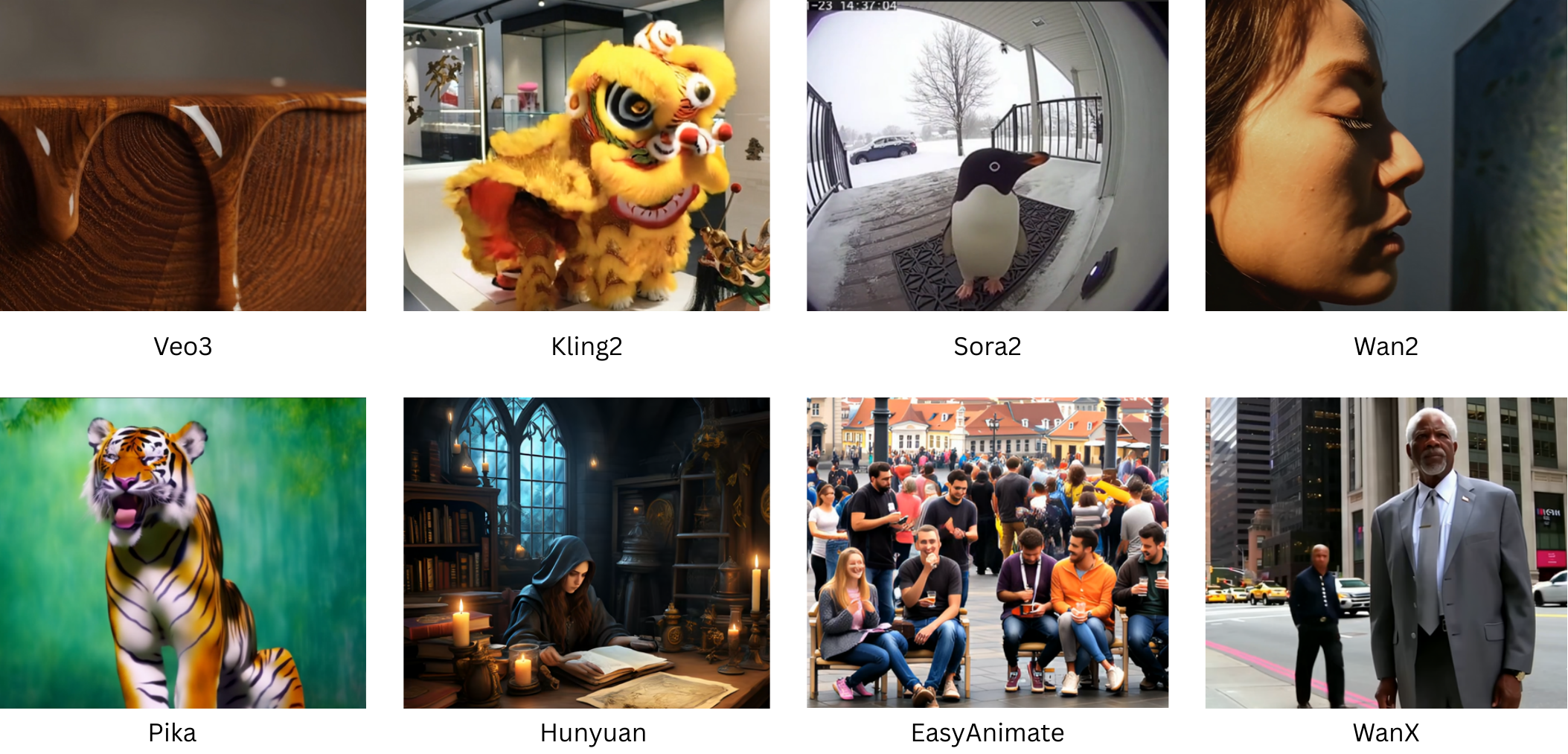}
  \caption{\textbf{Sampled video frames from some AI video generators.} Top: Recent generators produce high-quality visuals and realistic motion, closely resembling real videos. Bottom: Earlier models show clear artifacts, distorted content, and unnatural motion. }
  \label{fig:video_sample}
  \vspace{-0.6cm}
\end{figure}

As a result, reliable detection methods for AI-generated videos have become critically important. However, most prior work on synthetic media detection has focused on deepfakes \cite{DBLP:conf/cvpr/ZhaoZ0WZY21,yan2024df}, particularly face-centric manipulations, or AI-generated images \cite{chen2025mathcalxdfd, wen2025busterxmllmpoweredaigeneratedvideo,xu2025fakeshield}, which do not adequately capture the characteristics of fully generated videos produced by modern foundation models. To the best of our knowledge, only a small number of studies explicitly addressed AI-generated video detection before 2025, and the rapid emergence of high-quality text-to-video models has since exposed significant limitations in existing approaches. This growing gap highlights the urgent need for more robust, generalizable detection methods tailored to contemporary AI-generated video content.

Although AI-generated video detection has attracted increasing attention, existing approaches still face fundamental limitations stemming from both methodology and data. Some works \cite{corvi2025seeing,zhang2025physicsdriven,interno2025aigenerated} rely on physics-inspired or geometric priors. While conceptually appealing, these handcrafted assumptions often fail to generalize to modern high-fidelity generators, and their reported gains are frequently attributable to strong pretrained backbones rather than the proposed mechanisms. Other approaches \cite{Zheng_2025_D3, yan2025orthogonal_effort,Chen_2025_forgelen} adapt image-based detectors to videos, which is inherently limited because video synthesis introduces temporal dynamics and long-range dependencies that cannot be captured by frame-level analysis alone. MLLM-based methods \cite{wen2025busterxmllmpoweredaigeneratedvideo,park2025vidguardr1aigeneratedvideodetection,song2024on, li2025skyraaigeneratedvideodetection} offer flexibility but remain computationally expensive and unsuitable for large-scale deployment, while primarily relying on semantic reasoning rather than modeling the generative process itself. On the data side, existing benchmarks \cite{genvidbench,wang2024vidprom} are often constrained by outdated generators or limited coverage of recent commercial models, leading to insufficient generator diversity and weak cross-distribution evaluation.

Meanwhile, rapid advances in foundation video generation have fundamentally changed the detection landscape. Early detectors operated in pixel space, where visible artifacts provided reliable forensic cues \cite{ma2025detectingaigeneratedvideoframe_decof, corvi2025waverep}. Modern generators, however, are explicitly optimized to minimize pixel-level artifacts through diffusion models, transformers, and post-processing, making such cues increasingly unreliable. This shift suggests that detection must move beyond pixels and operate in representation space, where pretrained video encoders capture higher-level spatiotemporal dynamics that remain difficult for generative models to reproduce. While synthetic videos can achieve strong visual realism, they still struggle to match the temporal consistency and representation dynamics of real videos, motivating a representation-level paradigm for AI-generated video detection.

To explore this direction, we propose EA-Swin, an embedding-agnostic spatiotemporal detection head that operates directly on frozen video embeddings from foundation encoders. By decoupling detection from pixel-level processing, our approach enables scalable and robust detection for rapidly evolving video generators.

Our contributions are summarized as follows:
\begin{enumerate}
    \item We introduce \textbf{EA-Swin}, an embedding-agnostic spatiotemporal detection framework that operates directly on frozen video representations, shifting AI-generated video detection from pixel space to representation space through a factorized Swin-style transformer that models temporal dynamics and spatial coherence in embedding space while remaining compatible with generic ViT-style encoders.
    \item We construct EA-Video, a dataset of nearly 130K videos spanning commercial and open-source generators, and guarantee an unseen-generator protocol for cross-distribution evaluation.
    \item Extensive experiments demonstrate consistent improvements over prior methods on both seen and unseen generators, validating representation-level spatiotemporal modeling as a robust solution for modern AI-generated video detection.
\end{enumerate}

\section{Related Work}

\subsection{AI-generated video detection.} 
Recent years have seen a growing body of work on AI-generated video detection, surpassing earlier studies that mainly focused on deepfake face manipulation or image-level synthetic content. Early works such as DeCoF \cite{ma2025detectingaigeneratedvideoframe_decof} and DeMamba \cite{demamba} represent some of the first attempts to explicitly address general AI-generated video detection, highlighting the need to model temporal artifacts beyond static visual cues. Existing approaches can be categorized into video-based spatiotemporal models, embedding-trajectory–based methods, MLLM-based approaches, and image-based detectors commonly used for benchmarking.

\textbf{Video-based spatiotemporal models} aim to directly process video clips and capture temporal inconsistencies across frames. UNITE \cite{Kundu_2025_CVPR_UNITE} and DUB3D \cite{ji2024dub3d} employ 3D or video-level architectures to learn spatiotemporal representations, while DeCoF \cite{ma2025detectingaigeneratedvideoframe_decof} focuses on frequency-domain inconsistencies across time, and DeMamba \cite{demamba} introduces a structured state-space module to model local spatiotemporal irregularities. However, UNITE \cite{Kundu_2025_CVPR_UNITE} and DU3DB \cite{ji2024dub3d} are not open-sourced, and their video processing pipelines remain relatively coarse, often relying on short clips and limited temporal reasoning. Moreover, their evaluation protocols primarily use early or outdated generators, limiting their relevance to modern video synthesis models. Although DeMamba \cite{demamba} improves generalization through local spatiotemporal modeling, subsequent studies \cite{zhang2025physicsdriven_nsgvd,interno2025aigenerated_restrav} have shown that its performance can be surpassed when evaluated on newer, higher-quality generators, indicating limited robustness under rapidly evolving generation distributions.

\textbf{Embedding-based methods} analyze the temporal evolution of video representations extracted by pretrained encoders. Methods such as D3 measure simple differences between frame-level embeddings, while ResTraV \cite{interno2025aigenerated_restrav} and NSG-VD \cite{zhang2025physicsdriven_nsgvd} model higher-order temporal trajectories using statistics such as velocity, acceleration, or non-stationary graph structures; WaveRep \cite{corvi2025waverep} further augments this paradigm by analyzing frequency-domain dynamics of embedding sequences. Despite their conceptual simplicity and efficiency, these methods face intrinsic limitations as video generators improve: embeddings from real and synthetic videos increasingly overlap in representation space, weakening trajectory separability. In particular, simpler methods like D3 \cite{Zheng_2025_D3} become ineffective under modern generators, while ResTraV-style approaches (often relying on shallow MLP heads) lack sufficient capacity to capture deeper temporal dependencies, limiting their discriminative power and scalability.

More details on the AI-generated video detection method using image-based \cite{univd23,fredect20,cnnspot20,gramnet20,Chen_2025_forgelen,yan2025orthogonal_effort,npr24} or MLLM-based \cite{wen2025busterxmllmpoweredaigeneratedvideo,wen2026busterxunifiedcrossmodalaigenerated, li2025skyraaigeneratedvideodetection, park2025vidguardr1aigeneratedvideodetection, song2024_mm_det, fu2025learninghumanperceivedfakenessaigenerated, xiang2025aigvetoolaigeneratedvideoevaluation} methods are further discussed in the \textbf{Supplementary Material}.

\subsection{Benchmark datasets} 

The rapid evolution of video generation models makes constructing stable benchmarks for AI-generated video detection particularly challenging. Large-scale data collection is costly and time-consuming for open-source generators and financially expensive for commercial models. As a result, benchmarks such as VidProM \cite{wang2024vidprom} and GenVidBench \cite{genvidbench}, despite their scale, often become outdated within months as generation quality improves. Earlier generators like VideoCrafter2 \cite{chen2024videocrafter2}, Text2Video-Zero \cite{Khachatryan_2023_ICCV}, and MuseV \cite{xia2024musev} in these dataset quickly lose relevance because their artifacts are easily detected. Although later benchmarks such as RobustSora\cite{wang2025robustsora} and AIGVDBench\cite{ma2026onestopsolutionaigeneratedvideo} incorporate more recent models, they face the same issue of rapid obsolescence. Consequently, many recent studies construct task-specific datasets tailored to their methods and computational constraints (eg. GenBuster200k from BusterX \cite{wen2025busterxmllmpoweredaigeneratedvideo,wen2026busterxunifiedcrossmodalaigenerated}, Skyra \cite{li2025skyraaigeneratedvideodetection} introduces ViF-Bench, DeepTraceReward \cite{fu2025learninghumanperceivedfakenessaigenerated}), particularly for resource-intensive approaches such as MLLM-based models.

\section{EA-Video Dataset}
While recent datasets attempt to include newer video generators, they are limited by their small scale and reliance on generators similar to commercial models such as Sora2 or Veo3 due to cost constraints. To address the need for more diverse generators and datasets of larger scales, we introduce \textbf{EA-Video}. The construction of the EA-Video dataset are shown as below. 

\subsection{Dataset Curation}

First, for AI-generated videos, we leverage sources from previously published datasets. Video generators are selected based on the following criteria: 1) novelty of the generator; 2) generation quality (e.g., models that produce incoherent frames or meaningless content, such as T2VZ and MuseV, are excluded); 3) previously reported detection difficulty or accuracy \cite{ma2026onestopsolutionaigeneratedvideo,zhang2025physicsdriven_nsgvd,interno2025aigenerated_restrav}, excluding generators that are trivially distinguishable; 4) number of available videos per generator to ensure sufficient data; and 5) overall video quality, including prompt quality and video length.
We collect AI-generated videos from multiple sources, including videos from AIGVD \cite{ma2026onestopsolutionaigeneratedvideo}, VidProM \cite{wang2024vidprom}, GenBusterX \cite{wen2025busterxmllmpoweredaigeneratedvideo} \& GenBusterX++ \cite{wen2026busterxunifiedcrossmodalaigenerated}, ViF \cite{li2025skyraaigeneratedvideodetection}, DeepTraceReward \cite{fu2025learninghumanperceivedfakenessaigenerated}, and AIGVE \cite{xiang2025aigvetoolaigeneratedvideoevaluation}. To maintain dataset balance, when a generator produces an excessive number of videos, we cap its contribution to between 4k--7k videos.

In addition, we observe that many AI-generated videos are published on websites and social media platforms, making them a valuable data source. These videos are typically prompted by diverse users, are relatively long, and often undergo post-generation editing. To leverage this source, we collect AI-generated videos from publicly accessible platforms that provide video creation services. According to their descriptions, these platforms generate videos using pretrained models together with prompt engineering, post-processing, and fine-tuning strategies. We either create or obtain videos through OpenAI's Sora \cite{soraweb} and other platforms, including DigenAI \cite{digenai}, ImaStudio \cite{ima}, Invideo \cite{invideo}, OpenArt \cite{openart}, and Pollo AI \cite{pollo}.

\begin{figure}[!h]
  \centering
  \vspace{-0.5cm}
  \includegraphics[width=\textwidth]{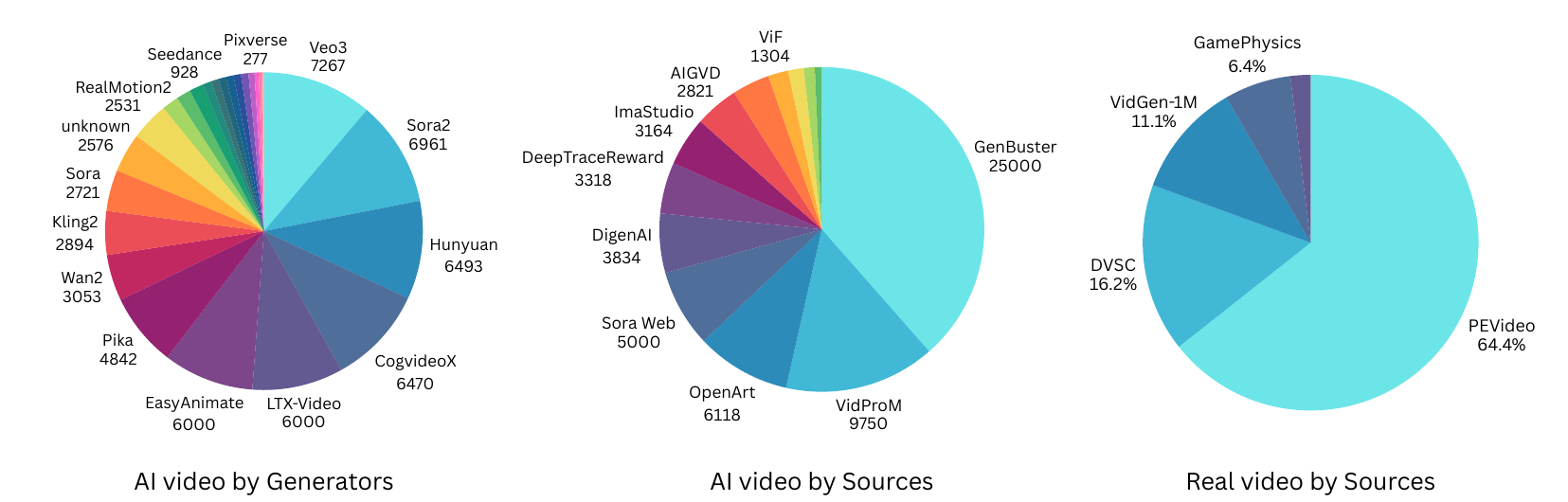}
  \caption{Real video data and AI videod data portion by Generators and Sources.
  }
  \label{fig:data_portion}
  \vspace{-0.5cm}
\end{figure}

Regarding real videos, we construct a dataset with a comparable scale and diverse sources. We primarily use videos from PEVideo \cite{bolya2025perception-encoder,cho2025perceptionlm} and further diversify the dataset with videos from DVSC \cite{dvsc}, VidGen-1M \cite{tan2024vidgen1mlargescaledatasettexttovideo}, GamePhysics \cite{gamephysics}, and VideoGameQA \cite{taesiri2025videogameqabench}. Notably, some video game datasets contain non-physical artifacts caused by in-game bugs; we include these videos in the dataset to examine potential confusion between such artifacts and AI-generated content.

After data collection, AI-generated videos are categorized by the generator. We then split the dataset into training, validation, and test sets. Specifically, generators with sufficient data (more than $3{,}000$ videos) are included in the training and validation sets, while generators with fewer samples are assigned to the test set, forming an unseen-generator benchmark. Real videos are split into training, validation, and test sets using the same ratios corresponding to each generator to ensure consistency across classes. More details about the dataset can be found in \textbf{Supplementary Material}.

\subsection{Dataset Composition}

\begin{wrapfigure}{r}{0.4\textwidth}
  \centering
  \vspace{-1cm}
  \includegraphics[width=0.4\textwidth]{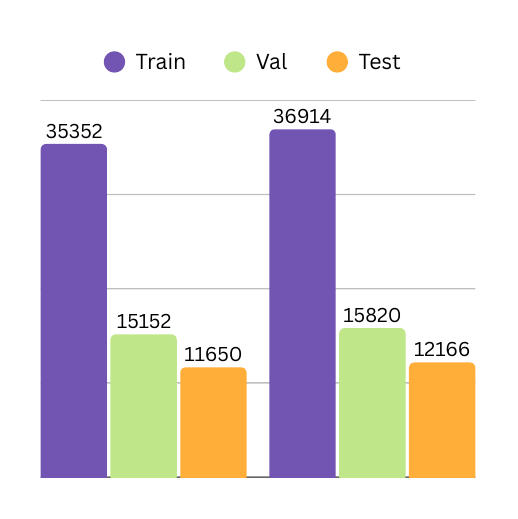}
  \vspace{-0.75cm}
  \caption{Train/Validation/Test set split.}
  \label{fig:data_split}
  \vspace{-0.5cm}
\end{wrapfigure}

As shown in Figure~\ref{fig:data_portion} and \ref{fig:data_split}, the final dataset comprises 127,054 videos, including approximately 65K AI-generated videos and more than 62K real videos. The data are balanced across training, validation, and test splits for both real and AI videos, with comparable proportions in each split. The figure further illustrates the distribution of videos by generators and data sources.

For AI-generated content, the dataset includes a large-scale collection from recent commercial and open-source generators, with strong representation from SoTA models as well as newer generators. The AI-generated content spans multiple generation tasks, including text-to-video, image-to-video, and video-to-video. The training and validation sets consist of videos generated by Veo3 \cite{veo3}, Sora2 \cite{sora2}, Hunyuan \cite{kong2024hunyuanvideo}, CogVideoX \cite{yang2025cogvideox}, EasyAnimate \cite{xu2024easyanimatehighperformancelongvideo}, LTX-Video \cite{ltx}, Pika \cite{pika}, Wan2 \cite{wan2025wanopenadvancedlargescale}, Kling2 \cite{klingteam2025klingomnitechnicalreport}, and Sora \cite{sora}. To evaluate generalization, the test set is composed of unseen generators in the train set, including RealMotion2 \cite{digenai}, Kling \cite{klingteam2025klingomnitechnicalreport}, Hailuo \cite{hailuo}, Seedance \cite{gao2025seedance10exploringboundaries}, Mochi \cite{genmo2024mochi}, Jimeng \cite{jimeng}, Gen3 \cite{gen3}, Luma \cite{luma}, Vidu \cite{bao2024viduhighlyconsistentdynamic}, Pyramids \cite{jin2025pyramidal}, SkyReels \cite{chen2025skyreelsv2infinitelengthfilmgenerative}, PixVerse \cite{pixverse}, Pika2 \cite{pika}, Gen4 \cite{gen4}, and an unknown category. The \textit{unknown} category contains videos whose exact generators were not disclosed by their creators and are primarily collected from GenBusterX++, ImaStudio, and InVideo. According to the disclosure of these platforms, these videos originate from a shared pool of recently released, high-quality generators such as Gen-3, Wan2, Veo3, and Kling2, and we therefore retain them as part of the unseen-generator test set. 

For real videos, the dataset is sourced mainly from PEVideo \cite{cho2025perceptionlm,bolya2025perception-encoder}, supplemented by DVSC \cite{dvsc}, VidGen-1M \cite{tan2024vidgen1mlargescaledatasettexttovideo}, VideoGameQA \cite{taesiri2025videogameqabench}, and GamePhysics \cite{gamephysics}, providing diverse real-world and synthetic-like artifacts.

\section{Method}
\label{sec:approach}
\subsection{Representation Trajectory Analysis}
\label{subsec:tsne_analysis}

To understand how real and AI-generated videos differ in representation space, 
we project frame-level embeddings from a pretrained video encoder into 2D using t-SNE and visualize their temporal trajectories (8 frames per video). 
Each polyline corresponds to the embedding evolution of a single video across time.

\begin{wrapfigure}{r}{0.6\textwidth}
  \centering
  \includegraphics[width=0.6\textwidth]{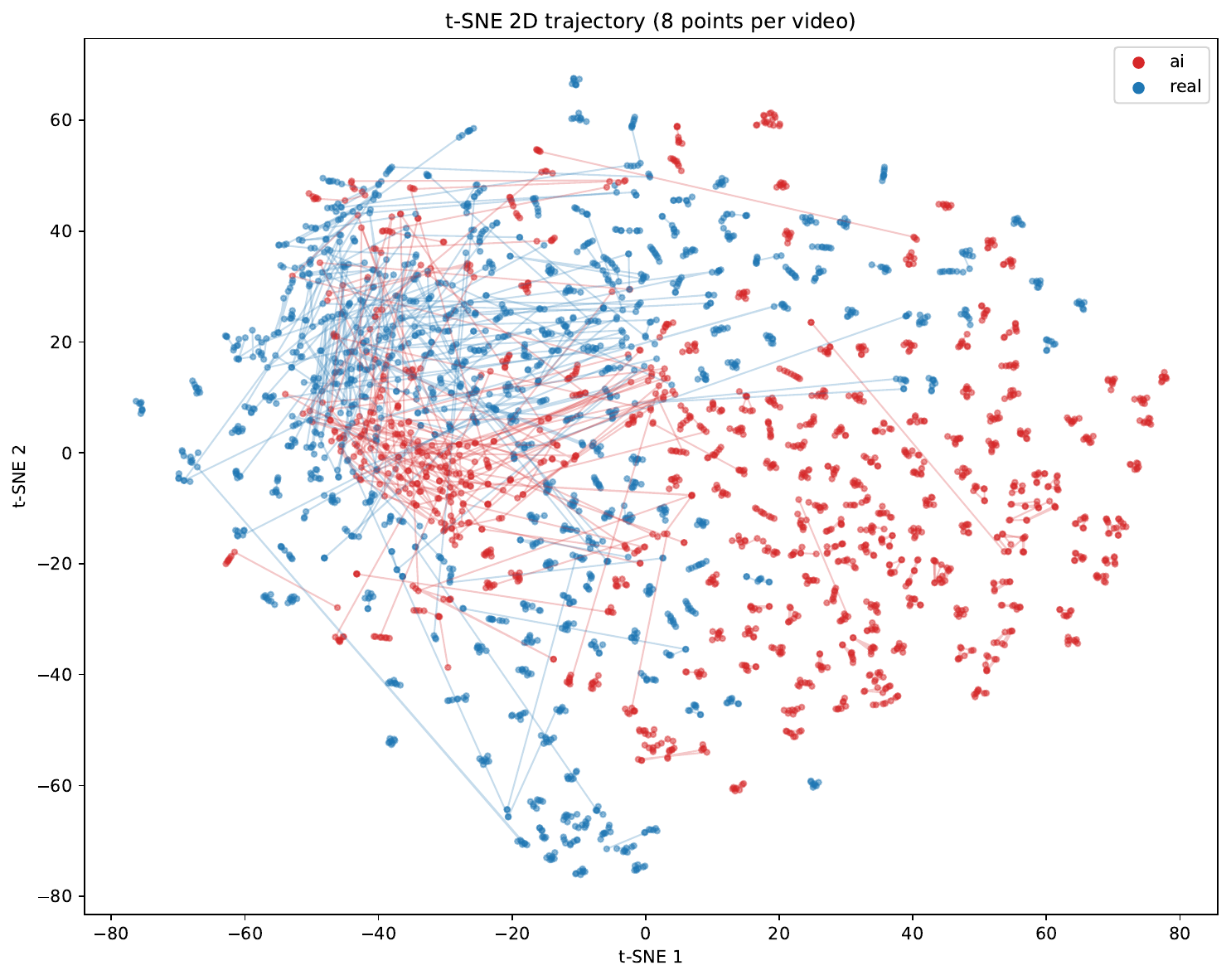}
  \caption{\textbf{t-SNE visualization of embedding trajectories.} Each polyline represents the temporal evolution of video embeddings from V-JEPA 2 encoder.}
  \label{fig:t-sne}
  \vspace{-0.7cm}
\end{wrapfigure}

As shown in Figure~\ref{fig:t-sne}, real and AI-generated videos partially overlap at early frames but gradually diverge as temporal dynamics unfold. While real videos exhibit diverse and irregular trajectory patterns, AI-generated videos tend to drift toward more concentrated regions with smoother and more constrained transitions. This suggests that although modern generators can closely match pixel-level appearance, they fail to fully reproduce the spatiotemporal dynamics captured by pretrained video representations. 

These observations indicate that temporal evolution in embedding space provides a stronger forensic signal than static frame-level analysis, motivating a detection framework that explicitly models representation trajectories rather than raw pixels.

\subsection{Embedding-Agnostic Spatiotemporal Modeling}

Based on the above analysis, we design a lightweight spatiotemporal detection head that operates directly on frozen video embeddings. 
Unlike Video Swin \cite{liu2022videoswin}, which processes high-dimensional pixel inputs using large spatial windows as a full video backbone, our setting operates on compact pretrained embeddings that already encode rich semantic and motion information. 
Therefore, instead of re-learning visual representations from pixels, we focus on modeling the temporal evolution and spatial coherence of embedding trajectories.

To this end, we propose \textbf{EA-Swin} (Embedding-Agnostic Swin Transformer), a factorized Swin-style transformer that performs temporal and spatial attention in embedding space. 
By decoupling detection from pixel-level processing, EA-Swin remains computationally efficient, encoder-agnostic, and specifically tailored for AI-generated video forensics.


 \begin{figure}[!h]
  \centering
  \includegraphics[width=\textwidth]{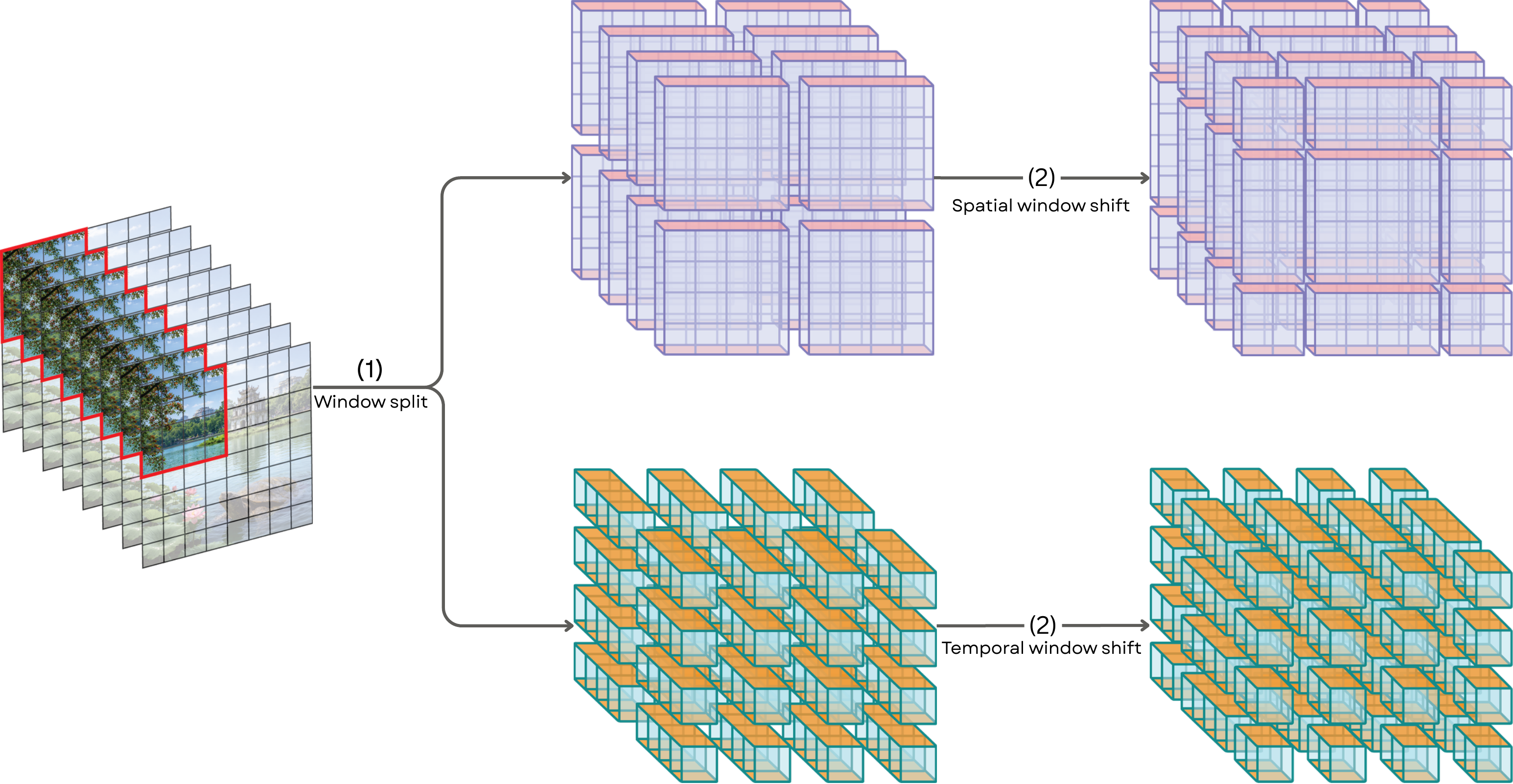}
  \caption{\textbf{Spatiotemporal window shifting mechanism.} The input video embedding is first partitioned into non-overlapping local windows (1) along spatial and temporal dimensions. To enable cross-window interaction and enhance global context modeling, the windows are then shifted (2) spatially across adjacent regions and temporally across neighboring frames.}
  \label{fig:window_shift}
  \vspace{-0.7cm}
\end{figure}

\subsection{Embedding-Agnostic Spatiotemporal Detection Head}
\label{sec:ea_swin}

\paragraph{Video embedding representation.}
Given a video $V$, we uniformly sample $T$ frames and extract features using a frozen pretrained video encoder. 
Depending on the backbone, embeddings may be frame-level or token-level. 
In the general case, each frame is decomposed into $S$ spatial tokens with dimension $D_{\text{in}}$, yielding a 4D representation
\[
\mathbf{Z} \in \mathbb{R}^{B \times T \times S \times D_{\text{in}}},
\]

where $B$ denotes the batch size. For frame-level encoders we set $S=1$. 
Since pretrained encoders already capture rich semantic and motion information, our goal is not to relearn visual features from pixels but to model the \emph{spatiotemporal evolution} of embedding trajectories.

\begin{wrapfigure}{r}{0.6\textwidth}
\vspace{-0.3cm}
  \centering
  \includegraphics[width=0.6\textwidth]{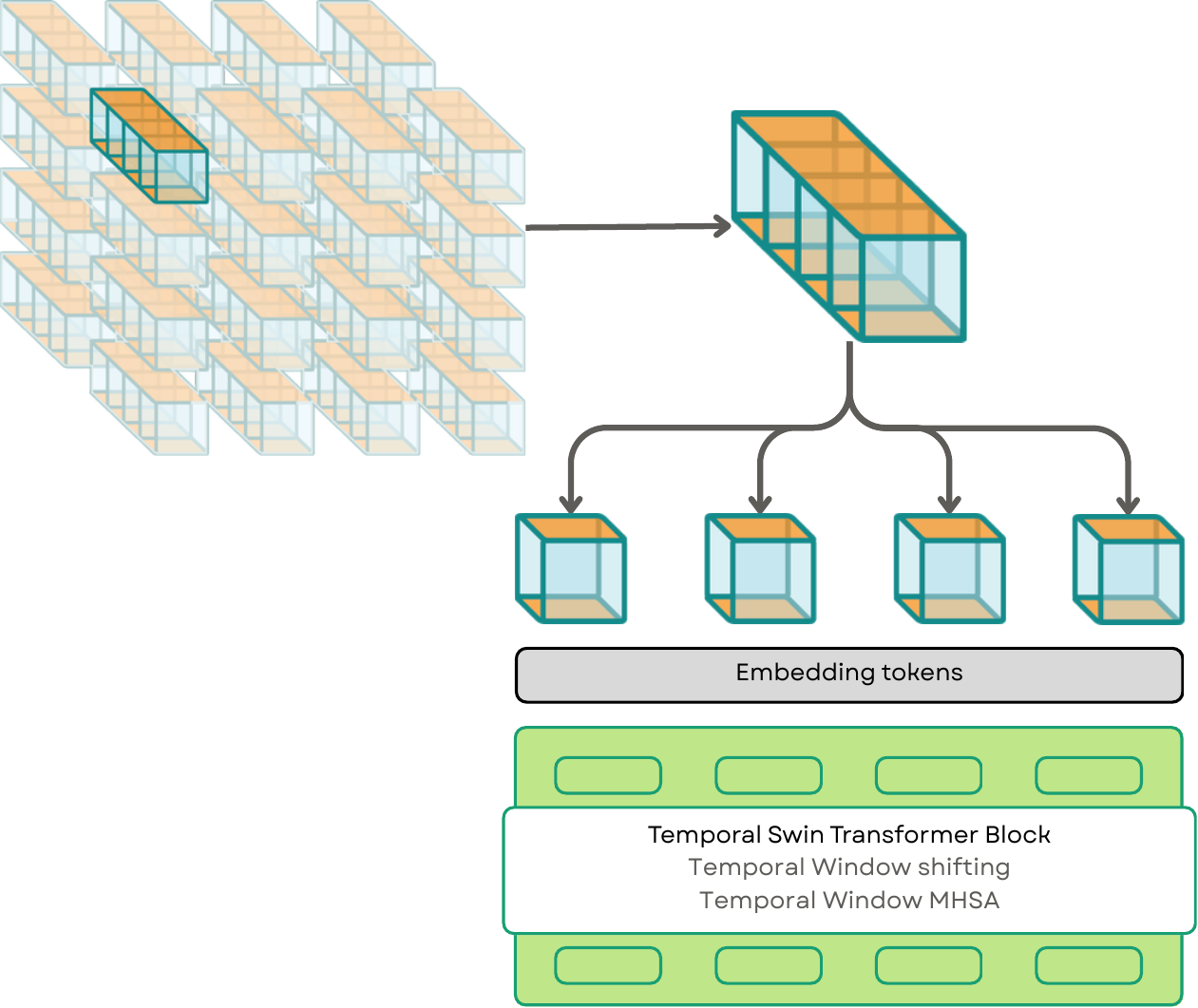}
  \caption{Temporal Swin attention.}
  \label{fig:temp_swin}
  \vspace{-1cm}
\end{wrapfigure}

\paragraph{Factorized spatiotemporal Swin attention.}
To model representation dynamics efficiently, we design a lightweight Swin-style detection head that alternates temporal and spatial windowed attention (Fig.~\ref{fig:window_shift}). 
Instead of applying joint attention over all $T\times S$ tokens, this factorized design significantly reduces computational cost while preserving long-range modeling capability. 
The window shifting mechanism enables information exchange across neighboring frames and spatial regions without incurring quadratic complexity.

\paragraph{Temporal Swin attention.}

\begin{wrapfigure}{r}{0.57\textwidth}
\vspace{-0.6cm}
  \centering
  \includegraphics[width=0.55\textwidth]{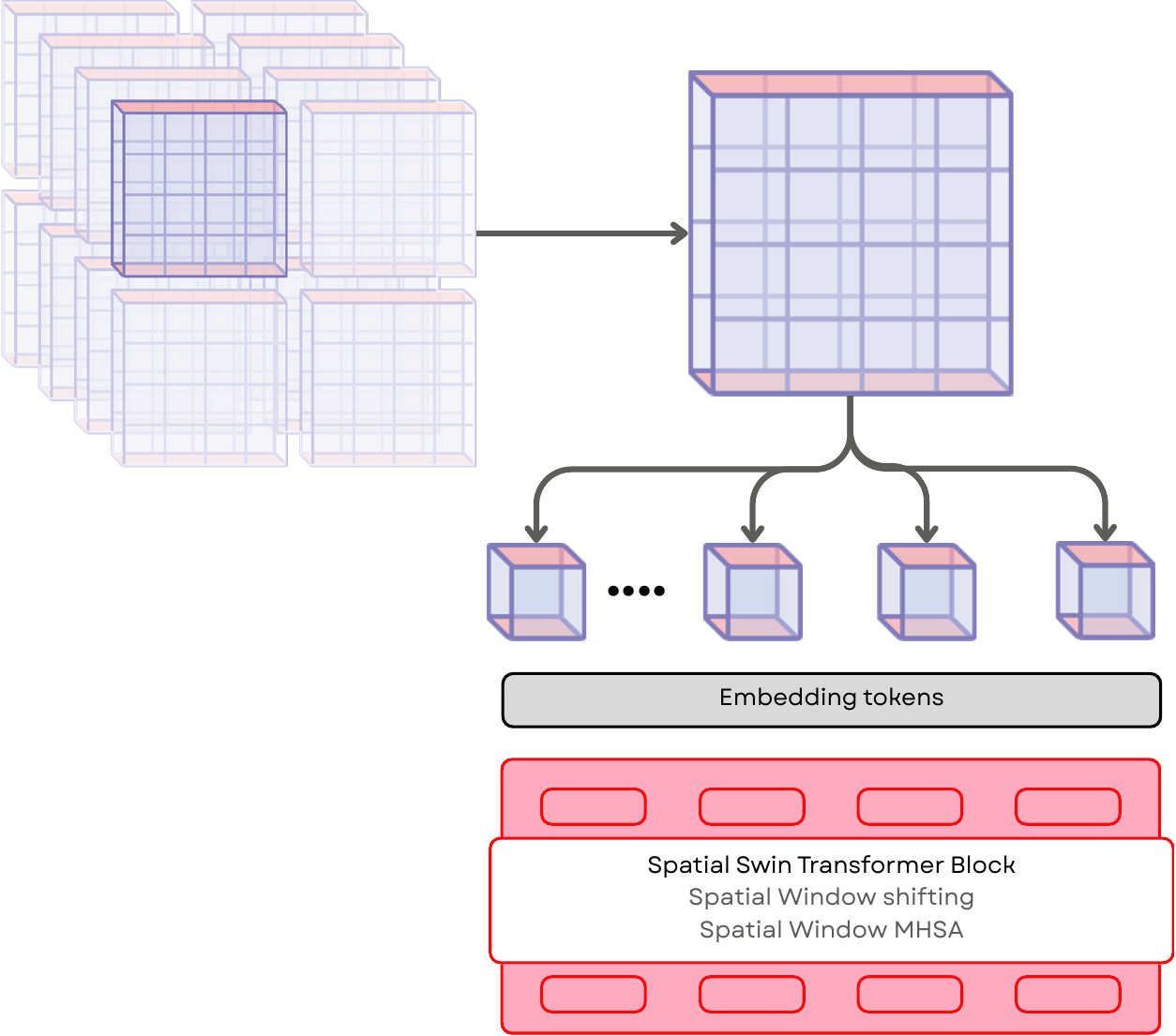}
  \caption{Spatial Swin attention.}
  \label{fig:spatial_swin}
  \vspace{-1cm}
\end{wrapfigure}

We first model temporal dependencies independently for each spatial token (Fig.~\ref{fig:temp_swin}) by reshaping
\[
\mathbf{Z}_t \in \mathbb{R}^{(B\cdot S)\times T \times D}.
\]
Windowed multi-head self-attention with window size $W_t$ is applied:
\[
\mathrm{Attn}_t(\mathbf{z})=
\mathrm{softmax}\!\left(\frac{QK^\top}{\sqrt{d_h}}+\mathbf{B}^{(t)}\right)V,
\]
where $\mathbf{B}^{(t)}$ is a learnable temporal relative positional bias. 
Alternating shifted windows allow cross-frame interaction while maintaining linear complexity in $T$.

\paragraph{Spatial Swin attention.}

After temporal modeling, spatial interactions within each frame are captured by reshaping tokens into a grid
\[
\mathbf{Z}_s \in \mathbb{R}^{(B\cdot T)\times H_p \times W_p \times D}.
\]
We then apply 2D windowed attention
\[
\mathrm{Attn}_s(\mathbf{z})=
\mathrm{softmax}\!\left(\frac{QK^\top}{\sqrt{d_h}}+\mathbf{B}^{(s)}\right)V,
\]
where $\mathbf{B}^{(s)}$ encodes spatial relative positional bias (Fig.~\ref{fig:spatial_swin}). 
Shifted windows again enable inter-window communication while preserving locality.

\begin{wrapfigure}{r}{0.37\textwidth}
  \centering
  \vspace{-0.5cm}
  \includegraphics[width=0.3\textwidth]{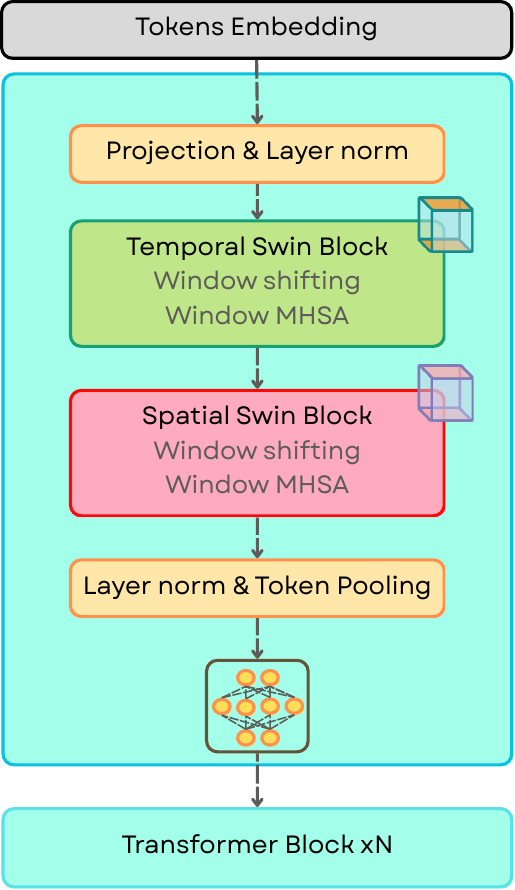}
  \caption{EA-Swin architecture.}
  \label{fig:ea_swin}
  \vspace{-0.7cm}
\end{wrapfigure}

\paragraph{Detection head and classification.}

The detection head consists of $D_t$ temporal blocks followed by $D_s$ spatial blocks (Fig.~\ref{fig:ea_swin}). 
Each block follows the standard transformer formulation
\[
\mathbf{y}=\mathbf{x}+\mathrm{MSA}(\mathrm{LN}(\mathbf{x})), \quad
\mathbf{z}=\mathbf{y}+\mathrm{MLP}(\mathrm{LN}(\mathbf{y})).
\]
After the final block, tokens are flattened and pooled to obtain a video-level representation, which is fed into a lightweight MLP classifier to predict whether the video is real or AI-generated.

\section{Experiment}

\subsection{Experimental details}
\textbf{Implementation.} We train a binary classifier (0: real, 1: AI-generated) using AdamW with learning rate $3\times10^{-4}$, weight decay $0.05$, cosine decay with 1 warmup epoch and minimum learning rate $10^{-6}$. Gradients are clipped to $1.0$ and automatic mixed precision (AMP) is enabled. All experiments are run with 3 random seeds on a single NVIDIA RTX 6000 Ada GPU (48GB).

The base AE-Swin uses a hidden size of $512$, $8$ attention heads, and V-JEPA2 as the vision encoder. Temporal and spatial window sizes are both set to $4$ with two transformer blocks for each. Every video produces $16$ embeddings; V-JEPA2 consumes $32$ frames via tubelets but outputs $16$ tokens, ensuring consistent inputs across experiments. More details of configuration are given in the \textbf{Supplementary Material}.

\textbf{Baselines \& Metrics.} We benchmark our method against recent SoTA approaches published in top-tier venues or widely adopted as strong baselines including DeMamba \cite{demamba}, NPR \cite{npr24}, STIL \cite{stil}, TALL \cite{tall}, ResTraV \cite{interno2025aigenerated_restrav}, D3 \cite{Zheng_2025_D3}, WaveRep Augmentation \cite{corvi2025waverep}, Forgelens \cite{Chen_2025_forgelen}, Effort \cite{yan2025orthogonal_effort}, and NSG-VD \cite{zhang2025physicsdriven_nsgvd}. We re-implement all the methods using our dataset;  only for WaveRep Augmentation, we used the pretrained weight since the training code is not released. For evaluation, we report Accuracy, Recall, F1-score, and AUC to comprehensively assess classification performance, robustness, and discrimination capability across different video generators.
\vspace{-0.5cm}

\subsection{Results}

\vspace{-0.25cm}

Table \ref{tab:bench_eval} reports the benchmark results on the seen generators. Traditional methods such as D3 struggle severely, performing close to random guessing (approximately 0.51 accuracy), highlighting the difficulty posed by modern high-quality video generators. More recent models including ResTraV, NPR, STIL, TALL, WaveRep, and DeMamba show progressively stronger performance, with DeMamba reaching 0.9515 average accuracy. Recently proposed detectors such as Forgelens and NSG-VD exhibit mixed results, with Forgelens achieving strong performance (0.977 accuracy) while NSG-VD remains unstable across generators. In contrast, EA-Swin consistently outperforms all baselines, achieving 0.9866 average accuracy, 0.9869 F1, and 0.9991 AUC, demonstrating both near-perfect discrimination capability and stable performance across diverse commercial generators. The consistent improvement across all metrics confirms the effectiveness of our spatiotemporal modeling design.

\begin{table}[!htbp]
\vspace{-0.6cm}
\centering
\caption{\textbf{Benchmark results on the evaluation (seen) set}, grouped by video generator. For each generator in the test set, the number of real videos is approximately balanced with the number of AI-generated videos from that generator. Abbreviation: HY (Hunyuan), CVX (CogVideoX), EA (EasyAnimate)}
\label{tab:bench_eval}
\resizebox{0.85\textwidth}{!}{%
\begin{tabular}{c|l|cccccccccc|
>{\columncolor[HTML]{DAE8FC}}c }
\hline
\multicolumn{1}{l|}{\textbf{Model}} & \textbf{Metric} & \textbf{Veo3} & \textbf{Sora2} & \textbf{HY} & \textbf{CVX} & \textbf{EA} & \textbf{LTX} & \textbf{Pika1} & \textbf{Wan2} & \textbf{Kling2} & \textbf{Sora} & \textbf{Avg} \\ \hline
 & Acc & 0.511 & 0.510 & 0.512 & 0.511 & 0.512 & 0.511 & 0.511 & 0.511 & 0.506 & 0.511 & 0.5105 \\
 & Recall & 1.000 & 1.000 & 1.000 & 1.000 & 1.000 & 1.000 & 1.000 & 0.999 & 0.993 & 1.000 & \textbf{0.9991} \\
 & F1 & 0.676 & 0.675 & 0.677 & 0.676 & 0.677 & 0.676 & 0.676 & 0.676 & 0.671 & 0.676 & 0.6757 \\
\multirow{-4}{*}{\textbf{D3}} & AUC & 0.374 & 0.671 & 0.190 & 0.304 & 0.168 & 0.366 & 0.264 & 0.573 & 0.548 & 0.319 & 0.3778 \\ \hline
 & Acc & 0.683 & 0.641 & 0.641 & 0.723 & 0.736 & 0.725 & 0.726 & 0.698 & 0.655 & 0.646 & 0.6874 \\
 & Recall & 0.890 & 0.802 & 0.986 & 0.956 & 0.988 & 0.963 & 0.968 & 0.903 & 0.810 & 0.837 & 0.9103 \\
 & F1 & 0.740 & 0.694 & 0.793 & 0.779 & 0.793 & 0.781 & 0.783 & 0.753 & 0.705 & 0.707 & 0.7528 \\
\multirow{-4}{*}{\textbf{ResTraV}} & AUC & 0.785 & 0.720 & 0.973 & 0.881 & 0.957 & 0.897 & 0.897 & 0.816 & 0.718 & 0.731 & 0.8375 \\ \hline
 & Acc & 0.841 & 0.636 & 0.847 & 0.847 & 0.844 & 0.857 & 0.839 & 0.722 & 0.606 & 0.834 & 0.7874 \\
 & Recall & 0.980 & 0.579 & 0.999 & 0.999 & 1.000 & 1.000 & 0.993 & 0.755 & 0.533 & 0.988 & 0.8825 \\
 & F1 & 0.863 & 0.618 & 0.869 & 0.870 & 0.868 & 0.878 & 0.863 & 0.735 & 0.579 & 0.859 & 0.8001 \\
\multirow{-4}{*}{\textbf{\begin{tabular}[c]{@{}c@{}}WaveRep\\ Augment\end{tabular}}} & AUC & 0.950 & 0.702 & 0.994 & 0.991 & 0.998 & 0.992 & 0.987 & 0.804 & 0.702 & 0.968 & 0.9089 \\ \hline
 & Acc & 0.945 & 0.944 & 0.948 & 0.948 & 0.957 & 0.956 & 0.956 & 0.955 & 0.956 & 0.952 & {\ul 0.9515} \\
 & Recall & 0.956 & 0.954 & 0.960 & 0.960 & 0.959 & 0.960 & 0.958 & 0.958 & 0.960 & 0.958 & 0.9581 \\
 & F1 & 0.954 & 0.954 & 0.955 & 0.958 & 0.957 & 0.956 & 0.956 & 0.955 & 0.956 & 0.953 & {\ul 0.9553} \\
\multirow{-4}{*}{\textbf{DeMamba}} & AUC & 0.960 & 0.960 & 0.960 & 0.960 & 0.960 & 0.960 & 0.960 & 0.960 & 0.960 & 0.960 & {\ul 0.9599} \\ \hline
 & Acc & 0.871 & 0.872 & 0.876 & 0.875 & 0.888 & 0.880 & 0.879 & 0.871 & 0.865 & 0.857 & 0.8734 \\
 & Recall & 0.917 & 0.922 & 0.935 & 0.930 & 0.940 & 0.930 & 0.931 & 0.921 & 0.929 & 0.902 & 0.9257 \\
 & F1 & 0.877 & 0.879 & 0.883 & 0.884 & 0.893 & 0.885 & 0.887 & 0.879 & 0.875 & 0.865 & 0.8807 \\
\multirow{-4}{*}{\textbf{NPR}} & AUC & 0.929 & 0.927 & 0.935 & 0.930 & 0.939 & 0.928 & 0.932 & 0.927 & 0.926 & 0.915 & 0.9288 \\ \hline
 & Acc & 0.784 & 0.833 & 0.860 & 0.815 & 0.895 & 0.807 & 0.825 & 0.824 & 0.820 & 0.696 & 0.8157 \\
 & Recall & 0.655 & 0.742 & 0.788 & 0.707 & 0.870 & 0.690 & 0.731 & 0.724 & 0.717 & 0.481 & 0.7104 \\
 & F1 & 0.764 & 0.826 & 0.856 & 0.803 & 0.897 & 0.793 & 0.817 & 0.815 & 0.809 & 0.630 & 0.8010 \\
\multirow{-4}{*}{\textbf{STIL}} & AUC & 0.894 & 0.918 & 0.939 & 0.924 & 0.952 & 0.915 & 0.927 & 0.916 & 0.920 & 0.863 & 0.9166 \\ \hline
 & Acc & 0.729 & 0.624 & 0.799 & 0.769 & 0.836 & 0.755 & 0.733 & 0.739 & 0.713 & 0.676 & 0.7372 \\
 & Recall & 0.661 & 0.472 & 0.801 & 0.735 & 0.875 & 0.716 & 0.683 & 0.691 & 0.641 & 0.560 & 0.6833 \\
 & F1 & 0.721 & 0.574 & 0.806 & 0.776 & 0.847 & 0.755 & 0.729 & 0.743 & 0.711 & 0.647 & 0.7308 \\
\multirow{-4}{*}{\textbf{TALL}} & AUC & 0.817 & 0.716 & 0.881 & 0.856 & 0.913 & 0.848 & 0.825 & 0.816 & 0.792 & 0.759 & 0.8224 \\ \hline
\multicolumn{1}{l|}{} & Acc & \multicolumn{1}{l}{0.966} & \multicolumn{1}{l}{0.964} & \multicolumn{1}{l}{0.985} & \multicolumn{1}{l}{0.983} & \multicolumn{1}{l}{0.994} & \multicolumn{1}{l}{0.986} & \multicolumn{1}{l}{0.986} & \multicolumn{1}{l}{0.964} & \multicolumn{1}{l}{0.982} & \multicolumn{1}{l|}{0.959} & \multicolumn{1}{r}{\cellcolor[HTML]{DAE8FC}0.977} \\
\multicolumn{1}{l|}{} & Recall & \multicolumn{1}{l}{0.941} & \multicolumn{1}{l}{0.940} & \multicolumn{1}{l}{0.980} & \multicolumn{1}{l}{0.975} & \multicolumn{1}{l}{0.999} & \multicolumn{1}{l}{0.984} & \multicolumn{1}{l}{0.977} & \multicolumn{1}{l}{0.938} & \multicolumn{1}{l}{0.970} & \multicolumn{1}{l|}{0.928} & \multicolumn{1}{r}{\cellcolor[HTML]{DAE8FC}0.963} \\
\multicolumn{1}{l|}{} & F1 & \multicolumn{1}{l}{0.966} & \multicolumn{1}{l}{0.964} & \multicolumn{1}{l}{0.985} & \multicolumn{1}{l}{0.983} & \multicolumn{1}{l}{0.995} & \multicolumn{1}{l}{0.986} & \multicolumn{1}{l}{0.986} & \multicolumn{1}{l}{0.963} & \multicolumn{1}{l}{0.983} & \multicolumn{1}{l|}{0.959} & \multicolumn{1}{r}{\cellcolor[HTML]{DAE8FC}0.977} \\
\multicolumn{1}{l|}{\multirow{-4}{*}{\textbf{Forgelens}}} & AUC & \multicolumn{1}{l}{0.996} & \multicolumn{1}{l}{0.993} & \multicolumn{1}{l}{0.999} & \multicolumn{1}{l}{0.998} & \multicolumn{1}{l}{1.000} & \multicolumn{1}{l}{0.999} & \multicolumn{1}{l}{0.997} & \multicolumn{1}{l}{0.993} & \multicolumn{1}{l}{0.999} & \multicolumn{1}{l|}{0.996} & \multicolumn{1}{r}{\cellcolor[HTML]{DAE8FC}0.997} \\ \hline
\multicolumn{1}{l|}{} & Acc & \multicolumn{1}{l}{0.542} & \multicolumn{1}{l}{0.485} & \multicolumn{1}{l}{0.653} & \multicolumn{1}{l}{0.586} & \multicolumn{1}{l}{0.563} & \multicolumn{1}{l}{0.610} & \multicolumn{1}{l}{0.608} & \multicolumn{1}{l}{0.478} & \multicolumn{1}{l}{0.489} & \multicolumn{1}{l|}{0.576} & \multicolumn{1}{r}{\cellcolor[HTML]{DAE8FC}0.559} \\
\multicolumn{1}{l|}{} & Recall & \multicolumn{1}{l}{0.650} & \multicolumn{1}{l}{0.518} & \multicolumn{1}{l}{0.873} & \multicolumn{1}{l}{0.737} & \multicolumn{1}{l}{0.687} & \multicolumn{1}{l}{0.781} & \multicolumn{1}{l}{0.766} & \multicolumn{1}{l}{0.507} & \multicolumn{1}{l}{0.529} & \multicolumn{1}{l|}{0.713} & \multicolumn{1}{r}{\cellcolor[HTML]{DAE8FC}0.676} \\
\multicolumn{1}{l|}{} & F1 & \multicolumn{1}{l}{0.587} & \multicolumn{1}{l}{0.501} & \multicolumn{1}{l}{0.716} & \multicolumn{1}{l}{0.640} & \multicolumn{1}{l}{0.611} & \multicolumn{1}{l}{0.667} & \multicolumn{1}{l}{0.661} & \multicolumn{1}{l}{0.492} & \multicolumn{1}{l}{0.508} & \multicolumn{1}{l|}{0.626} & \multicolumn{1}{r}{\cellcolor[HTML]{DAE8FC}0.601} \\
\multicolumn{1}{l|}{\multirow{-4}{*}{\textbf{NSG-VD}}} & AUC & \multicolumn{1}{l}{0.578} & \multicolumn{1}{l}{0.489} & \multicolumn{1}{l}{0.801} & \multicolumn{1}{l}{0.644} & \multicolumn{1}{l}{0.631} & \multicolumn{1}{l}{0.697} & \multicolumn{1}{l}{0.702} & \multicolumn{1}{l}{0.459} & \multicolumn{1}{l}{0.476} & \multicolumn{1}{l|}{0.636} & \multicolumn{1}{r}{\cellcolor[HTML]{DAE8FC}0.611} \\ \hline
 & Acc & 0.984 & 0.982 & 0.989 & 0.986 & 0.991 & 0.987 & 0.989 & 0.985 & 0.988 & 0.985 & \cellcolor[HTML]{FFCE93}\textbf{0.9866} \\
 & Recall & 0.982 & 0.984 & 0.997 & 0.992 & 0.999 & 0.997 & 0.990 & 0.993 & 0.990 & 0.986 & \cellcolor[HTML]{FFCE93}0.9911 \\
 & F1 & 0.984 & 0.982 & 0.989 & 0.987 & 0.991 & 0.988 & 0.989 & 0.986 & 0.989 & 0.986 & \cellcolor[HTML]{FFCE93}\textbf{0.9869} \\
\multirow{-4}{*}{\textbf{\begin{tabular}[c]{@{}c@{}}Ours \\ EA-Swin\end{tabular}}} & AUC & 0.998 & 0.998 & 1.000 & 0.999 & 1.000 & 1.000 & 0.999 & 0.999 & 0.999 & 0.997 & \cellcolor[HTML]{FFCE93}\textbf{0.9991} \\ \hline
\end{tabular}%
}
\vspace{-0.5cm}
\end{table}

Table \ref{tab:bench_test} presents results on unseen generators to evaluate cross-distribution generalization. Several prior methods experience noticeable performance degradation, most prominently WaveRep, which collapses on SKR, PV, Pika2, and Gen4 (e.g., 0.503/0.539/0.418/0.389 accuracy), and TALL, which also drops substantially on these generators. DeMamba remains the strongest baseline with an average accuracy of 0.922 and 0.948 AUC, while Forgelens shows high overall scores (0.882 accuracy, 0.971 AUC) but still exhibits instability on challenging cases such as Gen4. In contrast, EA-Swin demonstrates robust generalization, achieving 0.974 average accuracy and 0.997 AUC, while maintaining high Recall (Avg Recall 0.965) across nearly all unseen generators. These results suggest that EA-Swin captures more transferable generative artifacts rather than overfitting to the training distribution, delivering SoTA performance on both seen and emerging video generation models.

\begin{table}[]
\vspace{-0.5cm}
\centering
    \caption{\textbf{Benchmark results on the test (unseen) set}, grouped by video generator. Abbreviation: RM2 (Realmotion2), SD (SeeDance), JM (Jimeng), PRM (PyramidFlow), SKR (SkyReels), PV (PixVerse)}
\label{tab:bench_test}
\resizebox{\textwidth}{!}{%
\begin{tabular}{l|l|ccccccccccccccc|
>{\columncolor[HTML]{DAE8FC}}l }
\hline
\textbf{Model} & \textbf{Metric} & \textbf{Unk} & \textbf{RM2} & \textbf{Kling} & \textbf{Hailuo} & \textbf{SD} & \textbf{Mochi} & \textbf{JM} & \textbf{Gen3} & \textbf{Luma} & \textbf{Vidu} & \multicolumn{1}{l}{\textbf{PRM}} & \multicolumn{1}{l}{\textbf{SKR}} & \multicolumn{1}{l}{\textbf{PV}} & \multicolumn{1}{l}{\textbf{Pika2}} & \multicolumn{1}{l|}{\textbf{Gen4}} & \multicolumn{1}{c}{\cellcolor[HTML]{DAE8FC}\textbf{Avg}} \\ \hline
 & Acc & 0.511 & 0.511 & 0.512 & 0.510 & 0.512 & 0.515 & 0.512 & 0.512 & 0.513 & 0.512 & 0.511 & 0.519 & 0.515 & 0.514 & 0.545 & 0.515 \\
 & Recall & 1.000 & 1.000 & 1.000 & 0.997 & 1.000 & 0.998 & 0.998 & 1.000 & 1.000 & 1.000 & 1.000 & 0.992 & 1.000 & 1.000 & 0.981 & \textbf{0.998} \\
 & F1 & 0.676 & 0.676 & 0.677 & 0.675 & 0.677 & 0.678 & 0.676 & 0.677 & 0.678 & 0.677 & 0.676 & 0.678 & 0.678 & 0.678 & 0.688 & 0.678 \\
\multirow{-4}{*}{\textbf{D3}} & AUC & 0.507 & 0.321 & 0.281 & 0.572 & 0.449 & 0.467 & 0.293 & 0.389 & 0.246 & 0.347 & 0.290 & 0.411 & 0.447 & 0.544 & 0.550 & 0.408 \\ \hline
 & Acc & 0.628 & 0.699 & 0.718 & 0.654 & 0.619 & 0.638 & 0.703 & 0.599 & 0.726 & 0.698 & 0.713 & 0.722 & 0.701 & 0.676 & 0.701 & 0.680 \\
 & Recall & 0.774 & 0.901 & 0.928 & 0.791 & 0.741 & 0.822 & 0.940 & 0.696 & 0.956 & 0.925 & 0.943 & 0.925 & 0.917 & 0.914 & 0.883 & 0.870 \\
 & F1 & 0.680 & 0.753 & 0.771 & 0.700 & 0.665 & 0.699 & 0.764 & 0.639 & 0.781 & 0.758 & 0.771 & 0.773 & 0.758 & 0.742 & 0.751 & 0.734 \\
\multirow{-4}{*}{\textbf{ResTraV}} & AUC & 0.693 & 0.798 & 0.827 & 0.703 & 0.662 & 0.699 & 0.830 & 0.607 & 0.867 & 0.826 & 0.842 & 0.849 & 0.836 & 0.799 & 0.791 & 0.775 \\ \hline
 & Acc & 0.722 & 0.816 & 0.856 & 0.741 & 0.601 & 0.819 & 0.844 & 0.842 & 0.845 & 0.840 & 0.868 & 0.503 & 0.539 & 0.418 & 0.389 & 0.709 \\
 & Recall & 0.748 & 0.932 & 1.000 & 0.788 & 0.509 & 0.945 & 1.000 & 0.992 & 0.998 & 0.967 & 1.000 & 0.318 & 0.390 & 0.177 & 0.084 & 0.723 \\
 & F1 & 0.733 & 0.838 & 0.877 & 0.756 & 0.566 & 0.842 & 0.868 & 0.865 & 0.868 & 0.861 & 0.886 & 0.396 & 0.464 & 0.237 & 0.124 & 0.679 \\
\multirow{-4}{*}{\textbf{\begin{tabular}[c]{@{}l@{}}WaveRep\\ Augment\end{tabular}}} & AUC & 0.798 & 0.944 & 0.992 & 0.807 & 0.675 & 0.917 & 0.997 & 0.990 & 0.990 & 0.977 & 0.998 & 0.583 & 0.624 & 0.520 & 0.406 & 0.815 \\ \hline
 & Acc & 0.800 & 0.957 & 0.922 & 0.952 & 0.952 & 0.790 & 0.956 & 0.958 & 0.957 & 0.949 & 0.958 & 0.820 & 0.956 & 0.952 & 0.957 & {\ul 0.922} \\
 & Recall & 0.780 & 0.960 & 0.897 & 0.948 & 0.953 & 0.760 & 0.958 & 0.960 & 0.956 & 0.946 & 0.958 & 0.800 & 0.953 & 0.955 & 0.954 & 0.916 \\
 & F1 & 0.800 & 0.957 & 0.923 & 0.953 & 0.952 & 0.780 & 0.956 & 0.958 & 0.957 & 0.950 & 0.958 & 0.820 & 0.956 & 0.952 & 0.957 & {\ul 0.922} \\
\multirow{-4}{*}{\textbf{DeMamba}} & AUC & 0.900 & 0.960 & 0.957 & 0.959 & 0.960 & 0.890 & 0.960 & 0.960 & 0.960 & 0.959 & 0.960 & 0.910 & 0.960 & 0.960 & 0.960 & {\ul 0.948} \\ \hline
\multicolumn{1}{c|}{} & Acc & 0.854 & 0.882 & 0.849 & 0.858 & 0.877 & 0.758 & 0.868 & 0.871 & 0.864 & 0.868 & 0.890 & 0.797 & 0.817 & 0.792 & 0.781 & 0.842 \\
\multicolumn{1}{c|}{} & Recall & 0.895 & 0.939 & 0.871 & 0.904 & 0.927 & 0.695 & 0.923 & 0.908 & 0.916 & 0.931 & 0.937 & 0.786 & 0.819 & 0.795 & 0.695 & 0.863 \\
\multicolumn{1}{c|}{} & F1 & 0.861 & 0.888 & 0.857 & 0.866 & 0.883 & 0.760 & 0.876 & 0.878 & 0.872 & 0.876 & 0.895 & 0.807 & 0.823 & 0.799 & 0.773 & 0.848 \\
\multicolumn{1}{c|}{\multirow{-4}{*}{\textbf{NPR}}} & AUC & 0.914 & 0.938 & 0.909 & 0.922 & 0.928 & 0.841 & 0.932 & 0.933 & 0.920 & 0.928 & 0.873 & 0.815 & 0.835 & 0.808 & 0.795 & 0.886 \\ \hline
\multicolumn{1}{c|}{} & Acc & 0.667 & 0.844 & 0.609 & 0.748 & 0.729 & 0.567 & 0.760 & 0.696 & 0.709 & 0.735 & 0.742 & 0.591 & 0.660 & 0.596 & 0.595 & 0.683 \\
\multicolumn{1}{c|}{} & Recall & 0.430 & 0.763 & 0.318 & 0.578 & 0.536 & 0.258 & 0.621 & 0.490 & 0.520 & 0.563 & 0.585 & 0.283 & 0.436 & 0.271 & 0.301 & 0.464 \\
\multicolumn{1}{c|}{} & F1 & 0.585 & 0.840 & 0.490 & 0.712 & 0.682 & 0.399 & 0.743 & 0.636 & 0.660 & 0.697 & 0.711 & 0.436 & 0.584 & 0.429 & 0.453 & 0.604 \\
\multicolumn{1}{c|}{\multirow{-4}{*}{\textbf{STIL}}} & AUC & 0.842 & 0.935 & 0.825 & 0.897 & 0.888 & 0.735 & 0.899 & 0.861 & 0.861 & 0.891 & 0.893 & 0.788 & 0.843 & 0.785 & 0.762 & 0.847 \\ \hline
\multicolumn{1}{c|}{} & Acc & 0.635 & 0.791 & 0.656 & 0.673 & 0.715 & 0.631 & 0.804 & 0.758 & 0.740 & 0.709 & 0.796 & 0.662 & 0.639 & 0.582 & 0.530 & 0.688 \\
\multicolumn{1}{c|}{} & Recall & 0.483 & 0.775 & 0.527 & 0.568 & 0.627 & 0.475 & 0.822 & 0.740 & 0.658 & 0.646 & 0.775 & 0.549 & 0.512 & 0.389 & 0.327 & 0.592 \\
\multicolumn{1}{c|}{} & F1 & 0.587 & 0.796 & 0.633 & 0.679 & 0.701 & 0.592 & 0.816 & 0.762 & 0.729 & 0.700 & 0.806 & 0.635 & 0.604 & 0.502 & 0.432 & 0.665 \\
\multicolumn{1}{c|}{\multirow{-4}{*}{\textbf{TALL}}} & AUC & 0.729 & 0.874 & 0.759 & 0.763 & 0.804 & 0.723 & 0.880 & 0.832 & 0.837 & 0.803 & 0.880 & 0.782 & 0.744 & 0.696 & 0.606 & 0.781 \\ \hline
 & Acc & \multicolumn{1}{l}{0.916} & \multicolumn{1}{l}{0.996} & \multicolumn{1}{l}{0.729} & \multicolumn{1}{l}{0.979} & \multicolumn{1}{l}{0.978} & \multicolumn{1}{l}{0.623} & \multicolumn{1}{l}{0.994} & \multicolumn{1}{l}{0.989} & \multicolumn{1}{l}{0.968} & \multicolumn{1}{l}{0.983} & \multicolumn{1}{l}{0.982} & \multicolumn{1}{l}{0.686} & \multicolumn{1}{l}{0.948} & \multicolumn{1}{l}{0.882} & \multicolumn{1}{l|}{0.575} & 0.882 \\
 & Recall & \multicolumn{1}{l}{0.843} & \multicolumn{1}{l}{1.000} & \multicolumn{1}{l}{0.474} & \multicolumn{1}{l}{0.967} & \multicolumn{1}{l}{0.974} & \multicolumn{1}{l}{0.278} & \multicolumn{1}{l}{0.998} & \multicolumn{1}{l}{0.982} & \multicolumn{1}{l}{0.946} & \multicolumn{1}{l}{0.976} & \multicolumn{1}{l}{0.977} & \multicolumn{1}{l}{0.399} & \multicolumn{1}{l}{0.899} & \multicolumn{1}{l}{0.774} & \multicolumn{1}{l|}{0.175} & 0.777 \\
 & F1 & \multicolumn{1}{l}{0.911} & \multicolumn{1}{l}{0.996} & \multicolumn{1}{l}{0.641} & \multicolumn{1}{l}{0.979} & \multicolumn{1}{l}{0.978} & \multicolumn{1}{l}{0.429} & \multicolumn{1}{l}{0.994} & \multicolumn{1}{l}{0.989} & \multicolumn{1}{l}{0.968} & \multicolumn{1}{l}{0.984} & \multicolumn{1}{l}{0.982} & \multicolumn{1}{l}{0.565} & \multicolumn{1}{l}{0.947} & \multicolumn{1}{l}{0.870} & \multicolumn{1}{l|}{0.297} & 0.835 \\
\multirow{-4}{*}{\textbf{Forgelens}} & AUC & \multicolumn{1}{l}{0.989} & \multicolumn{1}{l}{1.000} & \multicolumn{1}{l}{0.930} & \multicolumn{1}{l}{0.997} & \multicolumn{1}{l}{0.997} & \multicolumn{1}{l}{0.844} & \multicolumn{1}{l}{1.000} & \multicolumn{1}{l}{0.999} & \multicolumn{1}{l}{0.996} & \multicolumn{1}{l}{0.998} & \multicolumn{1}{l}{0.999} & \multicolumn{1}{l}{0.959} & \multicolumn{1}{l}{0.994} & \multicolumn{1}{l}{0.987} & \multicolumn{1}{l|}{0.883} & 0.971 \\ \hline
 & Acc & \multicolumn{1}{l}{0.510} & \multicolumn{1}{l}{0.534} & \multicolumn{1}{l}{0.555} & \multicolumn{1}{l}{0.465} & \multicolumn{1}{l}{0.504} & \multicolumn{1}{l}{0.589} & \multicolumn{1}{l}{0.588} & \multicolumn{1}{l}{0.550} & \multicolumn{1}{l}{0.624} & \multicolumn{1}{l}{0.530} & \multicolumn{1}{l}{0.557} & \multicolumn{1}{l}{0.568} & \multicolumn{1}{l}{0.521} & \multicolumn{1}{l}{0.532} & \multicolumn{1}{l|}{0.605} & \cellcolor[HTML]{DAE8FC}0.549 \\
 & Recall & \multicolumn{1}{l}{0.589} & \multicolumn{1}{l}{0.635} & \multicolumn{1}{l}{0.672} & \multicolumn{1}{l}{0.517} & \multicolumn{1}{l}{0.564} & \multicolumn{1}{l}{0.733} & \multicolumn{1}{l}{0.689} & \multicolumn{1}{l}{0.676} & \multicolumn{1}{l}{0.828} & \multicolumn{1}{l}{0.650} & \multicolumn{1}{l}{0.662} & \multicolumn{1}{l}{0.694} & \multicolumn{1}{l}{0.621} & \multicolumn{1}{l}{0.656} & \multicolumn{1}{l|}{0.734} & \cellcolor[HTML]{DAE8FC}0.661 \\
 & F1 & \multicolumn{1}{l}{0.545} & \multicolumn{1}{l}{0.576} & \multicolumn{1}{l}{0.601} & \multicolumn{1}{l}{0.491} & \multicolumn{1}{l}{0.532} & \multicolumn{1}{l}{0.640} & \multicolumn{1}{l}{0.625} & \multicolumn{1}{l}{0.599} & \multicolumn{1}{l}{0.687} & \multicolumn{1}{l}{0.579} & \multicolumn{1}{l}{0.599} & \multicolumn{1}{l}{0.616} & \multicolumn{1}{l}{0.563} & \multicolumn{1}{l}{0.580} & \multicolumn{1}{l|}{0.646} & \cellcolor[HTML]{DAE8FC}0.592 \\
\multirow{-4}{*}{\textbf{NSG-VD}} & AUC & \multicolumn{1}{l}{0.524} & \multicolumn{1}{l}{0.555} & \multicolumn{1}{l}{0.592} & \multicolumn{1}{l}{0.445} & \multicolumn{1}{l}{0.501} & \multicolumn{1}{l}{0.662} & \multicolumn{1}{l}{0.607} & \multicolumn{1}{l}{0.563} & \multicolumn{1}{l}{0.716} & \multicolumn{1}{l}{0.566} & \multicolumn{1}{l}{0.585} & \multicolumn{1}{l}{0.614} & \multicolumn{1}{l}{0.546} & \multicolumn{1}{l}{0.567} & \multicolumn{1}{l|}{0.676} & \cellcolor[HTML]{DAE8FC}0.581 \\ \hline
\multicolumn{1}{c|}{} & Acc & 0.961 & 0.987 & 0.959 & 0.980 & 0.983 & 0.920 & 0.992 & 0.985 & 0.970 & 0.979 & 0.988 & 0.976 & 0.982 & 0.975 & 0.967 & \cellcolor[HTML]{FFCE93}\textbf{0.974} \\
\multicolumn{1}{c|}{} & Recall & 0.943 & 0.989 & 0.936 & 0.977 & 0.981 & 0.855 & 0.998 & 0.990 & 0.952 & 0.992 & 0.990 & 0.972 & 0.993 & 0.968 & 0.942 & \cellcolor[HTML]{FFCE93}{\ul 0.965} \\
\multicolumn{1}{c|}{} & F1 & 0.961 & 0.987 & 0.959 & 0.980 & 0.983 & 0.916 & 0.992 & 0.985 & 0.970 & 0.980 & 0.989 & 0.976 & 0.982 & 0.976 & 0.967 & \cellcolor[HTML]{FFCE93}\textbf{0.974} \\
\multicolumn{1}{c|}{\multirow{-4}{*}{\textbf{\begin{tabular}[c]{@{}c@{}}Ours \\ EA-Swin\end{tabular}}}} & AUC & 0.990 & 0.999 & 0.995 & 0.997 & 0.999 & 0.990 & 0.999 & 0.999 & 0.996 & 0.999 & 0.999 & 0.995 & 0.999 & 0.999 & 0.998 & \cellcolor[HTML]{FFCE93}\textbf{0.997} \\ \hline
\end{tabular}%
}
\vspace{-0.5cm}
\end{table}

Among the baselines, we observe that the embedding-based statistical methods (ResTraV, D3, NSG-VD) show clear limitations. Specifically, ResTraV and NSG-VD suffer from an information bottleneck due to aggressive dimensionality reduction, while D3 collapses to near-random performance by predicting almost all videos as fake (recall equals 1), revealing the weakness of heuristic variance-based criteria under modern high-quality generators. Although WaveRep Augmentation achieves competitive results, it mainly provides a data augmentation strategy; with a large-scale dataset such as EA-Video, its relative advantage becomes less pronounced. Among the baselines, DeMamba performs the best and confirms the importance of structured spatiotemporal modeling, yet it relies on a relatively large and computationally heavy architecture. Several other models exhibit noticeable performance drops on unseen generators, suggesting limited robustness and reliance on generator-specific artifacts. 

\subsection{Ablation Study}
\textbf{Architecture ablation.} We systematically simplify the proposed architecture in 4 ways to evaluate the contribution of each component:

\begin{enumerate}

    \item \textbf{Ablation 1}: We disable shifted windows by setting the window shift to 0.  
    \item \textbf{Ablation 2}: We replace the proposed temporal–spectral factorized attention with joint attention by flattening  T×S tokens and applying global window attention.
    \item \textbf{Ablation 3}: We replace attention pooling with simple mean pooling.
    \item \textbf{Ablation 4}: We replace the transformer head entirely with an MLP baseline.
    
\end{enumerate}

\begin{figure}[!h]
  \vspace{-0.5cm}
  \centering
  \includegraphics[width=\textwidth]{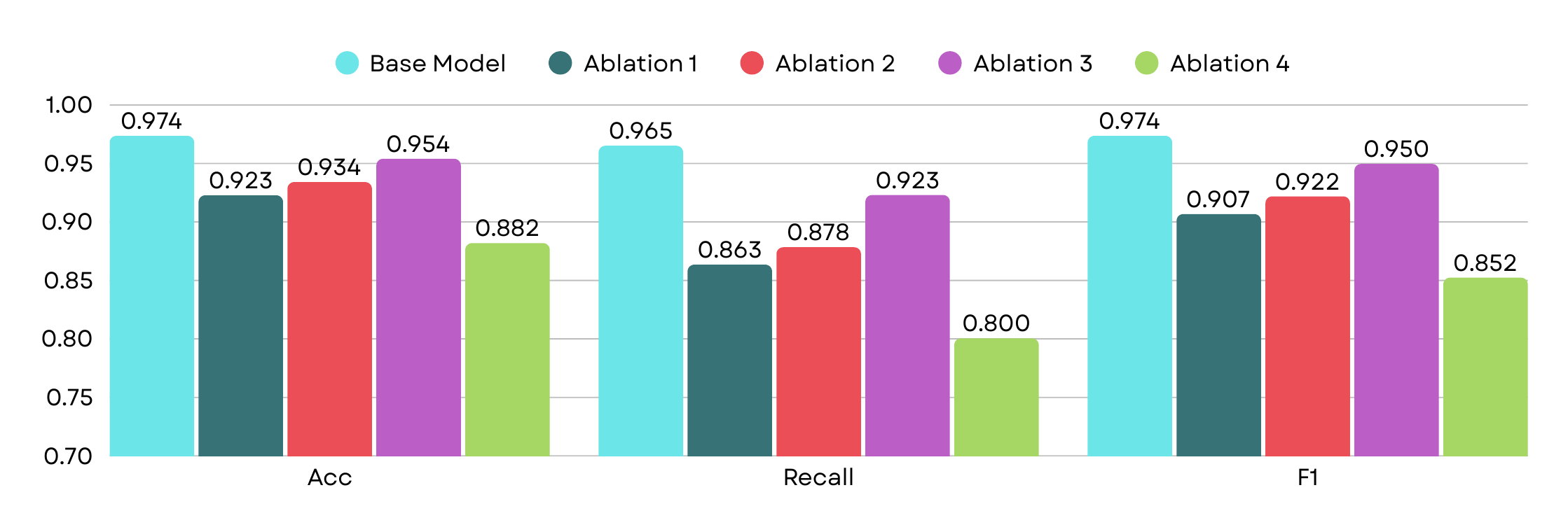}
  \vspace{-0.6cm}
  \caption{\text{Model architecture ablations experiment result on test set.} }
  \label{fig:arch_abl}
  \vspace{-0.6cm}
\end{figure}

The results (Figure \ref{fig:arch_abl}) confirm that each component of EA-Swin contributes meaningfully to performance. Removing shifted windows (Ablation 1) significantly reduces Recall, highlighting the importance of cross-window interaction, while replacing factorized temporal–spectral attention with joint attention (Ablation 2) leads to consistent degradation, showing the benefit of structured modeling. Substituting attention pooling with mean pooling (Ablation 3) further lowers performance, and the MLP baseline (Ablation 4) performs worst overall, especially in recall, demonstrating that both hierarchical attention and adaptive aggregation are crucial for robust detection.

\begin{figure}[!hbt]
\centering
\vspace{-0.5cm}
\begin{minipage}{0.64\textwidth}
\centering
\captionof{table}{Ablation on vision backbone}
\label{tab:abl_vis}
\resizebox{\textwidth}{!}{%
\begin{tabular}{@{}lcccccccccc@{}}
\toprule
\textbf{Backbone} & \multicolumn{5}{c}{\textbf{Val set}} & \multicolumn{5}{c}{\textbf{Test set}} \\
\cmidrule(l){2-6} \cmidrule(l){7-11}
 & Acc & Prec & Recall & F1 & AUC & Acc & Prec & Recall & F1 & AUC \\
\midrule
VJEPA2   & 0.986 & \textbf{0.986} & 0.987 & \textbf{0.987} & \textbf{0.999} & \textbf{0.975} & \textbf{0.985} & \textbf{0.966} & \textbf{0.975} & \textbf{0.997} \\
CLIP     & \textbf{0.987} & 0.983 & \textbf{0.991} & \textbf{0.987} & \textbf{0.999} & 0.974 & 0.983 & 0.965 & 0.974 & \textbf{0.997} \\
DINO3    & 0.971 & 0.976 & 0.968 & 0.972 & 0.995 & 0.891 & 0.970 & 0.811 & 0.865 & 0.970 \\
DINO2    & 0.954 & 0.956 & 0.953 & 0.955 & 0.990 & 0.874 & 0.949 & 0.791 & 0.846 & 0.957 \\
ConvNeXt2 & 0.977 & 0.980 & 0.975 & 0.978 & 0.997 & 0.916 & 0.976 & 0.858 & 0.913 & 0.976 \\
\bottomrule
\end{tabular}%
}
\end{minipage}
\hfill
\begin{minipage}{0.35\textwidth}
\caption{Video frames.}

\centering
\includegraphics[width=\textwidth]{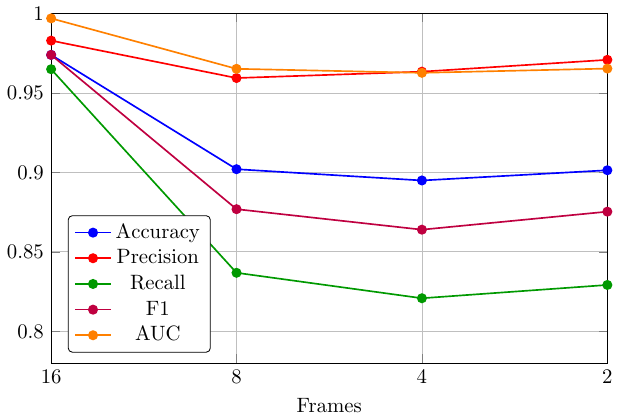}
\label{fig:video_frames}
\end{minipage}
\vspace{-0.7cm}
\end{figure}

\textbf{Vision encoder.} We evaluate the impact of different ViT-based vision backbones on our framework, including V-JEPA2 \cite{assran2025vjepa2selfsupervisedvideo}, CLIP \cite{CLIP}, DINOv3 \cite{simeoni2025dinov3}, and DINOv2 \cite{dinov2}, and a ViT-like encoder: ConvNeXt-v2 \cite{convnextv2}. As shown in Table \ref{tab:abl_vis}, V-JEPA2 consistently achieves the best performance on both validation and test sets, attaining the highest Accuracy, F1-score, and AUC, while CLIP remains competitive but slightly inferior. DINOv3 and DINOv2 show noticeably lower results, particularly on the test set where DINOv2 suffers the largest drop, indicating weaker generalization. Overall, these results suggest that stronger self-supervised spatiotemporal representations, as learned by V-JEPA2, provide more discriminative features for AI-generated video detection.

\vspace{-0.3cm}

\subsection{Robustness Test}

\textbf{Number of input frames.} We show the impact of reducing the number of input frames on model performance. As shown in Figure \ref{fig:video_frames}, decreasing the frame count from 16 to 8, 4, and 2 leads to a gradual decline across all metrics, with Recall and F1 being more sensitive to frame reduction. Nevertheless, the performance drop remains moderate, indicating that EA-Swin maintains reasonable robustness even under limited temporal information, while a higher number of frames still provides more stable and discriminative representations.

\begin{wraptable}{r}{0.42\textwidth}
\vspace{-0.7cm}
\centering
\caption{Robustness test for EA-Swin on the validation set}
\label{tab:robust}
\resizebox{0.40\textwidth}{!}{%
\begin{tabular}{@{}lcccc@{}}
\toprule
 & \textbf{Base} & \textbf{Blur} & \textbf{Comp.} & \textbf{Noise} \\ \midrule
\textbf{Acc} & 0.974 & 0.955 & 0.931 & 0.916 \\
\textbf{Prec} & 0.983 & 0.969 & 0.976 & 0.994 \\
\textbf{Recall} & 0.965 & 0.942 & 0.938 & 0.841 \\
\textbf{F1} & 0.974 & 0.955 & 0.956 & 0.912 \\
\textbf{AUC} & 0.997 & 0.991 & 0.990 & 0.988 \\ \bottomrule
\end{tabular}}
\vspace{-0.5cm}
\end{wraptable}

\textbf{Robustness Evaluation.} To evaluate robustness to common real-world video post-processing, we generate three validation variants using ffmpeg: H.264 re-encoding for compression (CRF 36), Gaussian noise with optional downscaling and temporal–uniform noise injection (CRF 40), and Gaussian blur ($\sigma=2$). Such perturbations commonly appear in videos shared on social media platforms due to re-encoding, resizing, and transmission artifacts. As shown in Table~\ref{tab:robust}, EA-Swin maintains stable performance across all perturbations with only moderate degradation from the clean setting (Acc 0.974, AUC 0.997). Under blur and compression, accuracy remains above 0.93 and AUC stays around 0.99, indicating strong resilience to realistic re-encoding artifacts. Gaussian noise is the most challenging condition, where accuracy drops to 0.916 and recall to 0.841, yet AUC remains high at 0.988. Overall, these results demonstrate consistent robustness to common video degradations.

\section{Conclusion}

We presented EA-Swin, an embedding-agnostic spatiotemporal detection framework for AI-generated video detection. 
Our results demonstrate that modeling the dynamics of pretrained video representations provides strong and consistent improvements in detection over prior pixel-level and trajectory-based approaches. 
These findings suggest that modern AI-generated video detection should shift from pixel-space analysis toward representation-space modeling, where temporal consistency and higher-level structure remain difficult for generative models to reproduce.

More broadly, this work highlights the growing importance of representation-level forensics in the era of foundation video models. 
As generative systems continue to improve visual realism, detection methods must increasingly rely on higher-level spatiotemporal signals rather than visible artifacts. 
We hope that EA-Swin and the EA-Video benchmark will serve as a foundation for future research on scalable and robust synthetic video detection.



\section*{Acknowledgements}
We would like to express our gratitude to our colleagues at N2TP (Phong Ho, Nhung Duong, Trang Pham) for their assistance with the research. The primary author would like to thank Quang Hung Nguyen (Viettel) for his assistance in data collection.

%
%
\bibliographystyle{splncs04}
\bibliography{main}

\newpage

\appendix

\section{More Related Work}
\subsection{AI-Generated videos and emerging issues}
Recent years have witnessed rapid advances in video generation models. Video generators dating back to 2023 and early 2024 (e.g., VideoCrafter2 \cite{chen2024videocrafter2}, Text2Video-Zero \cite{Khachatryan_2023_ICCV}, ModelScope \cite{wang2023modelscopetexttovideotechnicalreport}) suffered from noticeable artifacts, temporal inconsistency, and unrealistic motion, rendering synthetic videos relatively easy to identify. However, the advent of commercial models such as Veo \cite{veo} and Sora \cite{sora} in mid-2024, followed by newer generations including Veo3 \cite{veo3}, Sora2 \cite{sora2}, Gen3 \cite{gen3}, Vidu \cite{bao2024viduhighlyconsistentdynamic}, and Kling \cite{klingteam2025klingomnitechnicalreport}, has significantly narrowed the perceptual gap between real and synthetic videos. In parallel, open-source models such as OpenSora \cite{opensora, opensora2}, Pyramid Flow \cite{jin2025pyramidal}, CogVideoX \cite{yang2025cogvideox}, and Wan \cite{wan2025wanopenadvancedlargescale} have rapidly undergone enhancements, enabling high-quality video synthesis and lowering the barrier to large-scale deployment.

Beyond technical progress, recent studies highlight increasing societal and security concerns surrounding AI-generated video content. Prior work shows that AI disclosure can influence user engagement and perceived quality, but its effectiveness depends on users’ trust in AI systems \cite{CHEN2025108448}. Other research emphasizes that the widespread adoption of generative AI has outpaced the development of effective safeguards, enabling malicious misuse such as fraud, misinformation, and large-scale deception \cite{YOON2025101491,easttom}. As synthetic videos become more realistic, disclosure and manual inspection become unreliable, motivating the need for robust, content-based video detection methods.

\subsection{More AI-generated video detection methods}

\textbf{Multimodal large language models (MLLMs).} Another line of work explores the use of MLLMs for AI-generated video detection: BusterX \cite{wen2025busterxmllmpoweredaigeneratedvideo,wen2026busterxunifiedcrossmodalaigenerated}, Skyra \cite{li2025skyraaigeneratedvideodetection}, Vidguard-R1 \cite{park2025vidguardr1aigeneratedvideodetection}, MM-Det \cite{song2024_mm_det}, DeepTraceReward \cite{fu2025learninghumanperceivedfakenessaigenerated}, AIGVE \cite{xiang2025aigvetoolaigeneratedvideoevaluation}. While these approaches benefit from strong semantic understanding and interpretability, they suffer from two key limitations. As MLLMs are typically large and highly general-purpose, making them computationally expensive and poorly suited for scalable video-level detection. Moreover, several studies indicate that such models often focus on describing video content rather than performing true forensic detection, effectively assessing whether the model can reason about or narrate potential artifacts instead of learning discriminative signals for real-versus-generated classification. As a result, MLLM-based approaches remain more aligned with video understanding or analysis tasks than robust, standalone video detection.

\textbf{Image-based detectors} such as UnivFD \cite{univd23}, Gram-Net \cite{gramnet20}, NPR \cite{npr24}, CNNSpot \cite{cnnspot20}, FreDect\cite{fredect20}, or more recently ForgeLens \cite{Chen_2025_forgelen} and Effort \cite{yan2025orthogonal_effort}, were originally designed for AI-generated image detection and are often repurposed for video by frame sampling and score aggregation. While these methods are useful for benchmarking and benefit from strong pretrained vision backbones, they fundamentally ignore temporal structure and long-range motion consistency. As a result, they struggle to distinguish high-quality AI-generated videos whose individual frames appear realistic, making them unsuitable as standalone solutions for video-level detection.

\textbf{Deepfake detection.} Recent advances in deepfake detection have focused on improving robustness, generalization, and interpretability under increasingly realistic generation techniques \cite{deepfakeeccv2,deepfakeeccv1,Hu_2025_ICCV}. Early approaches primarily relied on CNN-based classifiers and frequency-domain analysis to capture forgery artifacts, such as abnormal high-frequency patterns or spatial inconsistencies. More recent works leverage transformer architectures and spatiotemporal modeling to capture subtle temporal inconsistencies across frames \cite{11094369}. For example, AdvOU \cite{Li_2025_ICCV} introduces an adversarial framework to discover and mitigate unfairness and bias in deepfake detectors, improving reliability and cross-domain generalization. Other studies explore human-inspired contextual reasoning, such as HICOM \cite{Hu_2025_ICCV}, which incorporates scene motion coherence, inter-face consistency, and gaze alignment to improve detection in multi-face scenarios. Additionally, multimodal approaches have emerged to enhance detection performance and interpretability. For instance, recent vision-language frameworks formulate deepfake detection as a reasoning task, enabling models to leverage semantic and textual cues alongside visual features to improve generalization and provide interpretable explanations \cite{deepfakeeccv1}. These advances highlight the importance of modeling spatial, temporal, and semantic inconsistencies for robust deepfake detection, aligning closely with video understanding and spatiotemporal representation learning.

\subsection{Vision encoders}
Recent progress in representation learning has led to the emergence of large-scale vision encoders trained either through contrastive language supervision or purely self-supervised objectives. These encoders aim to produce transferable visual representations that generalize across tasks such as classification, detection, segmentation, video understanding, and even planning.

\textbf{Contrastive Vision–Language Pretraining} \cite{CLIP} introduced CLIP, a large-scale vision–language model trained on 400M image–text pairs using a contrastive objective. By aligning image and text embeddings in a shared space, CLIP enables strong zero-shot transfer to downstream classification tasks without task-specific fine-tuning. CLIP demonstrated that language supervision can serve as a scalable proxy for semantic labeling, establishing a new paradigm for foundation vision models.
Subsequent open reproductions such as OpenCLIP \cite{cherti2023reproducible} further scaled data and model sizes, improving robustness and cross-dataset generalization. However, vision–language pretraining is inherently constrained by the availability and quality of aligned image–text pairs, and textual supervision may not capture fine-grained spatial or low-level visual details.

\textbf{Self-Supervised Image Encoders}
DINO \cite{dino} and its successors demonstrated that self-distillation without labels can produce semantically meaningful visual features. Building upon this line of work, DINOv2 \cite{dinov2} scaled self-supervised training to curated large-scale datasets (142M images) and billion-parameter Vision Transformers. DINOv2 combined improvements in data curation, stabilization techniques, and distillation to produce robust, general-purpose features that rival or surpass supervised and vision–language counterparts on both image-level and pixel-level tasks.
More recently, DINOv3 \cite{simeoni2025dinov3} further explores scaling laws, architecture refinements, and training stabilization for foundation vision encoders, improving robustness, efficiency, and transfer across a broader distribution of tasks. These DINO-based models emphasize that carefully scaled discriminative self-supervision can produce foundation features without relying on language alignment.

\textbf{Joint-Embedding Predictive Architectures for Video}
Extending self-supervised learning to the temporal domain, Self-Supervised Learning from Video with a Joint-Embedding Predictive Architecture introduced V-JEPA \cite{vjepa} , a joint-embedding predictive architecture that learns by predicting masked spatio-temporal representations in a latent space rather than reconstructing pixels. By focusing on predictable aspects of the scene, JEPA-style training avoids modeling high-frequency details irrelevant to semantic understanding. Building on this approach, V-JEPA 2 \cite{assran2025vjepa2selfsupervisedvideo} scaled video pretraining to over one million hours of internet video. V-JEPA 2 demonstrates that large-scale action-free pretraining yields representations suitable for motion understanding, action anticipation, video question answering (after language alignment), and even downstream robotic planning when augmented with limited interaction data. These results suggest that predictive self-supervision in representation space can serve as a foundation for world models.

\textbf{Convolutional Modernization: ConvNeXt}
In parallel to transformer-based encoders, ConvNeXt \cite{liu2022convnet} convolutional architectures by modernizing ResNet designs with training strategies and architectural choices inspired by Vision Transformers. ConvNeXt demonstrated that pure convolutional networks, when appropriately scaled and regularized, remain competitive with transformer-based encoders. ConvNeXt V2 \cite{convnextv2} further integrates masked autoencoding into ConvNeXt pretraining, bridging convolutional inductive biases with self-supervised masked modeling objectives. This highlights that architectural choice and pretraining objective are deeply intertwined, and strong visual representations can emerge from both convolutional and transformer families.

\section{More detail on dataset}
\subsection{Dataset detail}

We provide detailed statistics of the dataset composition in Table~\ref{tab:dataset_detail}. The table reports the number of AI-generated videos per generator and split, along with the corresponding number of paired real videos used as negative samples. As discussed, the training and test splits share the same generator families, while the validation split is constructed from unseen generators to evaluate out-of-distribution generalization.

\begin{table}[]
\centering
\caption{Dataset composition per generator and split}
\label{tab:dataset_detail}
\resizebox{0.7\textwidth}{!}{%
\begin{tabular}{cccccc}
\hline
\textbf{Split} & \textbf{Generator} & \textbf{\#AI\_vids} & \textbf{\%in\_split} & \textbf{\#correspond\_real} & \textbf{Published} \\ 
\cmidrule(l){1-1} \cmidrule(l){2-2} \cmidrule(l){3-3} \cmidrule(l){4-4} \cmidrule(l){5-5} \cmidrule(l){6-6}
\multicolumn{6}{c}{\textbf{Train set}} \\ \hline
train & veo3 & 5054 & 13.691 & - & 7/2025 \\
train & sora2 & 4857 & 13.158 & - & 9/2025 \\
train & cogvideox & 4605 & 12.475 & - & 3/2025 \\
train & hunyuan & 4524 & 12.256 & - & 3/2025 \\
train & easyanimate & 4199 & 11.375 & - & 7/2024 \\
train & ltxvideo & 4130 & 11.188 & - & 12/2024 \\
train & pika1 & 3388 & 9.178 & - & 12/2023 \\
train & wan2 & 2155 & 5.838 & - & 4/2025 \\
train & kling2 & 2056 & 5.570 & - & 4/2025 \\
train & sora & 1917 & 5.193 & - & 2/2024 \\ \hline
\multicolumn{6}{c}{\textbf{Val set}} \\ \hline
val & veo3 & 2213 & 13.989 & 2119 & - \\
val & sora2 & 2104 & 13.300 & 2016 & - \\
val & hunyuan & 1969 & 12.446 & 1886 & - \\
val & ltxvideo & 1870 & 11.820 & 1791 & - \\
val & cogvideox & 1865 & 11.789 & 1788 & - \\
val & easyanimate & 1801 & 11.384 & 1725 & - \\
val & pika1 & 1454 & 9.191 & 1393 & - \\
val & wan2 & 898 & 5.676 & 859 & - \\
val & kling2 & 838 & 5.297 & 802 & - \\
val & sora & 804 & 5.082 & 771 & - \\ \hline
\multicolumn{6}{c}{\textbf{Test set}} \\ \hline
test & unknown & 2576 & 21.174 & 2467 & - \\
test & realmotion2 & 2531 & 20.804 & 2424 & 11/2024 \\
test & kling1 & 1106 & 9.091 & 1060 & 2024 \\
test & hailuo & 949 & 7.800 & 910 & 2024 \\
test & seedance & 928 & 7.628 & 888 & 6/2025 \\
test & mochi1 & 580 & 4.767 & 555 & 9/2025 \\
test & jimeng & 501 & 4.118 & 480 & 2025 \\
test & gen3 & 500 & 4.110 & 479 & 6/2024 \\
test & luma & 500 & 4.110 & 478 & 4/2025 \\
test & vidu & 491 & 4.036 & 469 & 5/2024 \\
test & pyramid & 488 & 4.011 & 468 & 3/2025 \\
test & skyreels & 399 & 3.280 & 382 & 4/2025 \\
test & pixverse & 277 & 2.277 & 265 & 2025 \\
test & pika2 & 186 & 1.529 & 178 & 2025 \\
test & gen4 & 154 & 1 & 147 & 3/2025 \\ \hline
\end{tabular}%
}
\end{table}

\subsection{Video frames from generators}

\begin{figure}[]
  \centering
  \includegraphics[width=\textwidth]{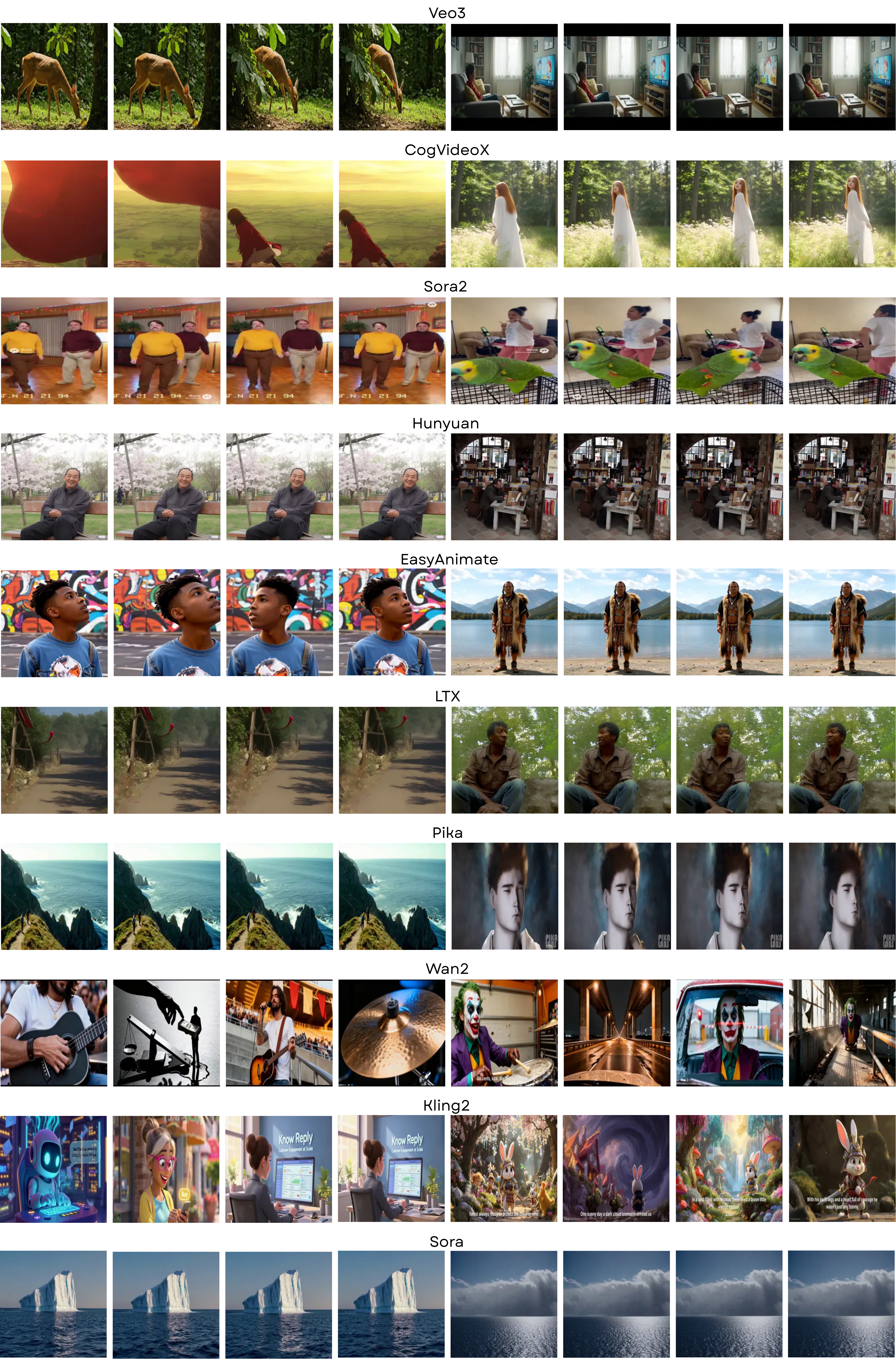}
  \label{fig:video_sample}
\end{figure}

\begin{figure}[]
  \centering
  \includegraphics[width=\textwidth]{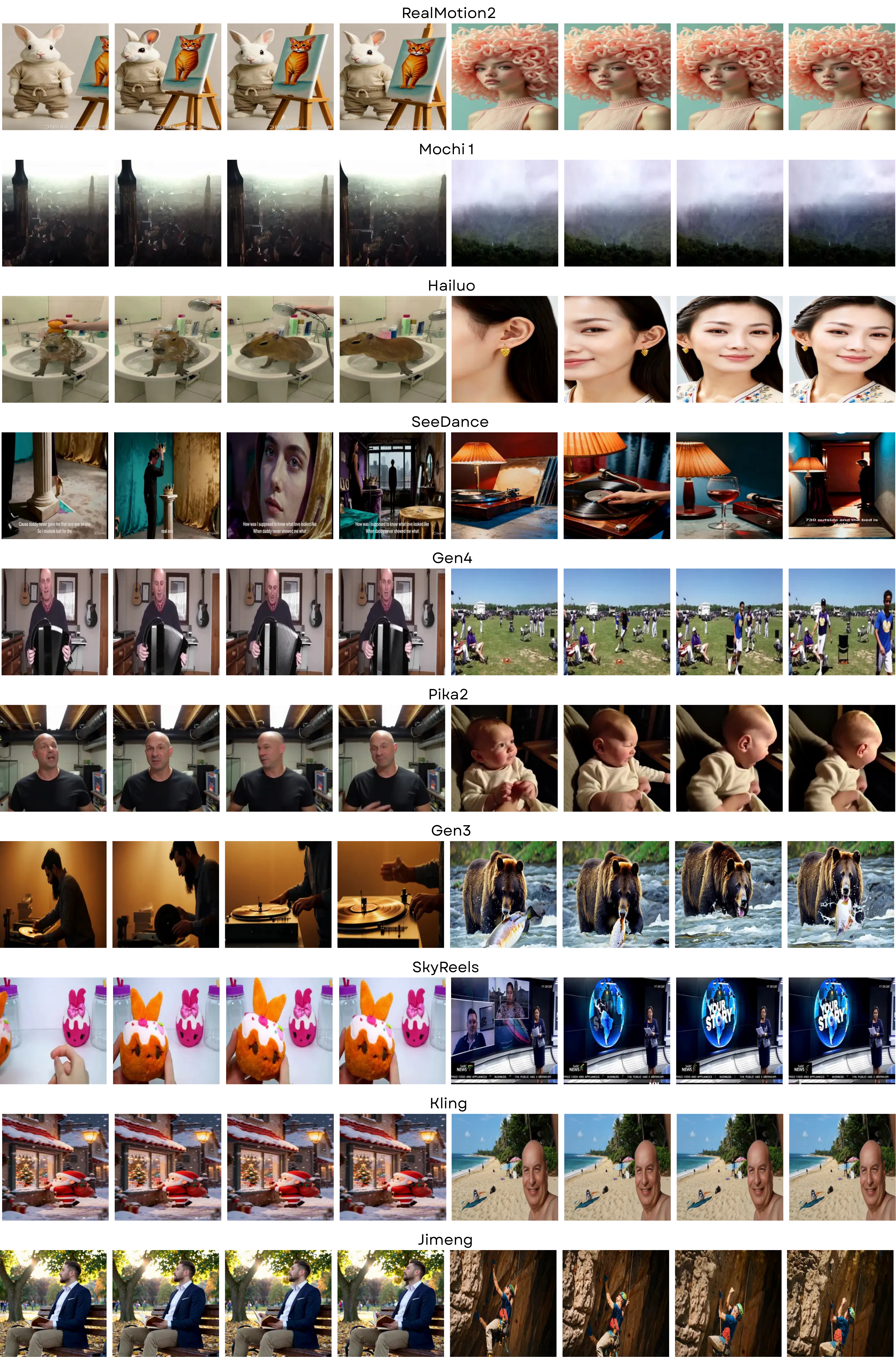}
  \label{fig:video_sample}
\end{figure}

\begin{figure}[]
  \centering
  \includegraphics[width=\textwidth]{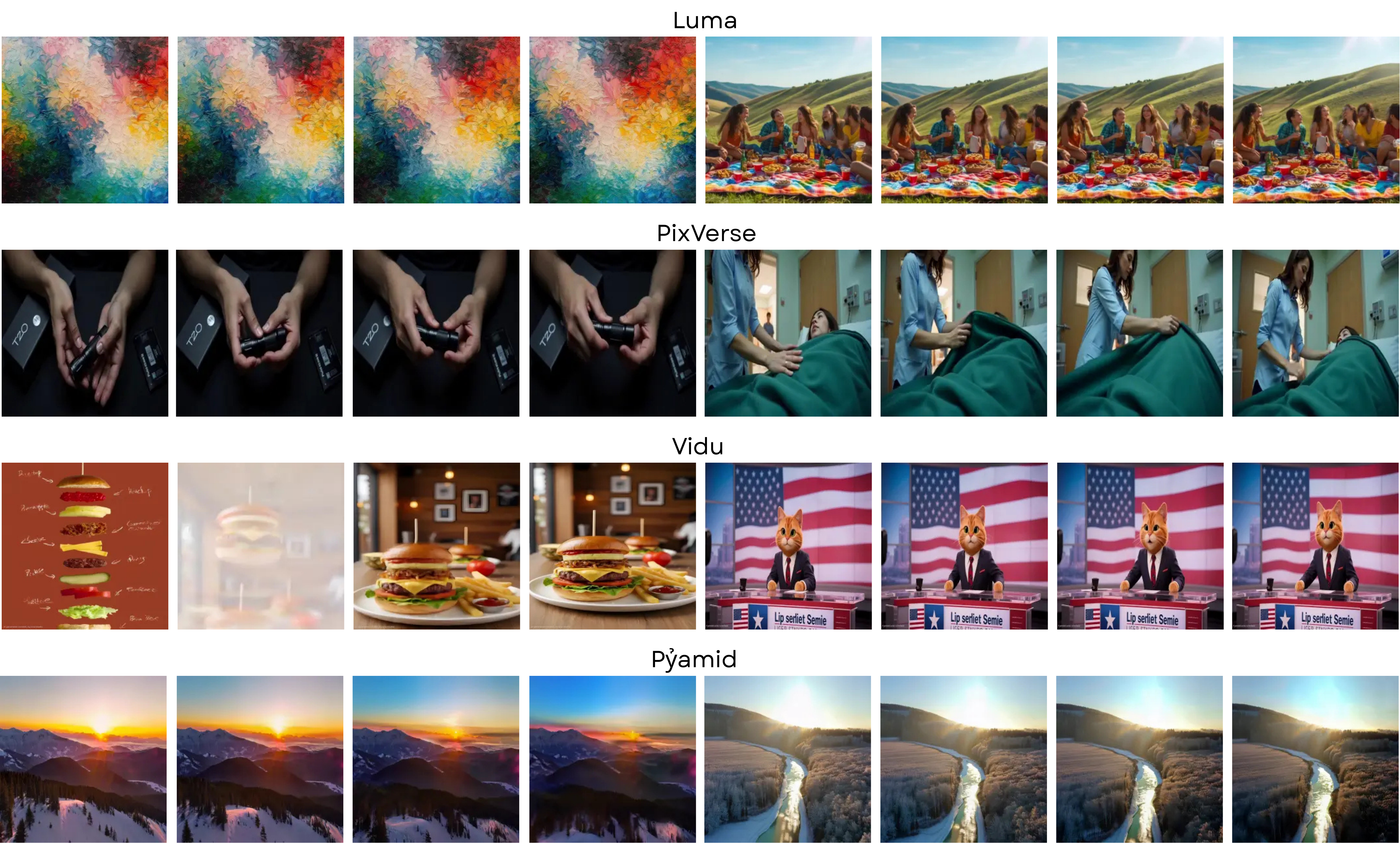}
  \label{fig:video_sample}
\end{figure}

\section{Deatil config \& Hardware}

\begin{table}[]
\centering
\caption{Base training config detail}
\label{tab:base_config}
\resizebox{0.75\textwidth}{!}{%
\begin{tabular}{lll}
\hline
\textbf{Category} & \textbf{Setting} & \textbf{Value} \\ \hline
\multirow{9}{*}{Optimizer} & Optimizer & AdamW \\
 & Learning rate & 3e-4 \\
 & Weight decay & 0.05 \\
 & LR schedule & Cosine decay \\
 & Warmup & 1 epoch \\
 & Min learning rate & 1e-6 \\
 & Gradient clipping & 1.0 \\
 & Mixed precision & AMP enabled \\
 & Random seeds & 3 \\ \hline
\multirow{7}{*}{Model (AE-Swin base)} & Hidden dimension & 512 \\
 & Attention heads & 8 \\
 & Vision encoder & V-JEPA2 \\
 & Temporal window size & 4 \\
 & Spatial window size & 4 \\
 & Temporal blocks & 2 \\
 & Spatial blocks & 2 \\ \hline
\multirow{2}{*}{Input processing} & Per video embeddings & 16 \\
 & Raw frames to V-JEPA2 & 32 (2-frame tubelets) \\ \hline
\multirow{3}{*}{Hardware} & GPU & RTX 6000 Ada (48GB) \\
 & VRAM used & est. 42GB \\
 & Training setup & Single GPU \\ \hline
\multirow{3}{*}{Disk Space} & Video & 355GB \\
 & Per embedding file (.pt) & 8.1MB \\
 & Embedding total & 1.1TB \\ \hline
\multirow{5}{*}{Vision Encoder} & V-JEPA 2 & vjepa2-vitl-fpc64-256 \\
 & CLIP & clip-vit-large-patch14 \\
 & DINO-v3 & dinov3-vitl16-pretrain \\
 & DINO-v2 & dinov2-base \\
 & ConvNeXt & convnextv2-large-22k-384 \\ \hline
\end{tabular}%
}
\end{table}

\section{Extended results}
Below, we present more experiment result. First, the hyperparameters sweep is shown in Tables \ref{tab:ps_val} \& \ref{tab:ps_test}. Second, we present the result of ablation study experimented on each vision encoders: V-JEPA 2 (Tables \ref{tab:vjp_val} \& \ref{tab:vjp_test}), CLIP (Tables \ref{tab:clip_val} \& \ref{tab:clip_val}), DINO-v3 (Tables \ref{tab:dino3_val} \& \ref{tab:dino3_test}), DINO-v2 (Tables \ref{tab:dino2_val} \& \ref{tab:dino2_test}).
\begin{table}[]
\centering
\caption{Hyperparameter sweep result on val set}
\label{tab:ps_val}
\resizebox{0.85\textwidth}{!}{%
\begin{tabular}{@{}ccccccccccccc@{}}
\toprule
\textbf{Model} & \textbf{Configs} & \textbf{Metric} & Veo3 & Sora2 & HY & CVX & EA & LTX & Pika1 & Wan2 & Kling2 & Sora \\ \midrule
\multicolumn{13}{c}{Model Dimension} \\ \midrule
\multicolumn{1}{c|}{\multirow{5}{*}{\begin{tabular}[c]{@{}c@{}}Dimension \\ reduction\end{tabular}}} & \multicolumn{1}{c|}{\multirow{5}{*}{\begin{tabular}[c]{@{}c@{}}--d\_model 256 \\ --depth\_t 2 \\ --depth\_s 2\end{tabular}}} & \multicolumn{1}{c|}{Acc} & 0.976 & 0.977 & 0.988 & 0.986 & 0.990 & 0.988 & 0.985 & 0.985 & 0.987 & 0.985 \\
\multicolumn{1}{c|}{} & \multicolumn{1}{c|}{} & \multicolumn{1}{c|}{Prec} & 0.986 & 0.984 & 0.983 & 0.986 & 0.984 & 0.982 & 0.986 & 0.981 & 0.988 & 0.988 \\
\multicolumn{1}{c|}{} & \multicolumn{1}{c|}{} & \multicolumn{1}{c|}{Recall} & 0.967 & 0.971 & 0.994 & 0.988 & 0.997 & 0.995 & 0.986 & 0.989 & 0.987 & 0.983 \\
\multicolumn{1}{c|}{} & \multicolumn{1}{c|}{} & \multicolumn{1}{c|}{F1} & 0.977 & 0.978 & 0.989 & 0.987 & 0.990 & 0.988 & 0.986 & 0.985 & 0.987 & 0.985 \\
\multicolumn{1}{c|}{} & \multicolumn{1}{c|}{} & \multicolumn{1}{c|}{AUC} & 0.997 & 0.997 & 0.999 & 0.999 & 1.000 & 1.000 & 0.999 & 0.999 & 0.999 & 0.998 \\ \midrule
\multicolumn{1}{c|}{\multirow{5}{*}{\begin{tabular}[c]{@{}c@{}}Dimension \\ increase\end{tabular}}} & \multicolumn{1}{c|}{\multirow{5}{*}{\begin{tabular}[c]{@{}c@{}}--d\_model 768 \\ --depth\_t 2 \\ --depth\_s 2\end{tabular}}} & \multicolumn{1}{c|}{Acc} & 0.984 & 0.982 & 0.989 & 0.986 & 0.991 & 0.987 & 0.989 & 0.985 & 0.988 & 0.985 \\
\multicolumn{1}{c|}{} & \multicolumn{1}{c|}{} & \multicolumn{1}{c|}{Prec} & 0.985 & 0.980 & 0.981 & 0.981 & 0.983 & 0.978 & 0.989 & 0.978 & 0.987 & 0.985 \\
\multicolumn{1}{c|}{} & \multicolumn{1}{c|}{} & \multicolumn{1}{c|}{Recall} & 0.982 & 0.984 & 0.997 & 0.992 & 0.999 & 0.997 & 0.990 & 0.993 & 0.990 & 0.986 \\
\multicolumn{1}{c|}{} & \multicolumn{1}{c|}{} & \multicolumn{1}{c|}{F1} & 0.984 & 0.982 & 0.989 & 0.987 & 0.991 & 0.988 & 0.989 & 0.986 & 0.989 & 0.986 \\
\multicolumn{1}{c|}{} & \multicolumn{1}{c|}{} & \multicolumn{1}{c|}{AUC} & 0.998 & 0.998 & 1.000 & 0.999 & 1.000 & 1.000 & 0.999 & 0.999 & 0.999 & 0.997 \\ \midrule
\multicolumn{13}{c}{Spatial Depth} \\ \midrule
\multicolumn{1}{c|}{\multirow{5}{*}{\begin{tabular}[c]{@{}c@{}}Spatial \\ Depth\\ decrease\end{tabular}}} & \multicolumn{1}{c|}{\multirow{5}{*}{\begin{tabular}[c]{@{}c@{}}--d\_model 512 \\ --depth\_t 2 \\ --depth\_s 1\end{tabular}}} & \multicolumn{1}{c|}{Acc} & 0.981 & 0.979 & 0.988 & 0.987 & 0.991 & 0.989 & 0.986 & 0.986 & 0.984 & 0.986 \\
\multicolumn{1}{c|}{} & \multicolumn{1}{c|}{} & \multicolumn{1}{c|}{Prec} & 0.984 & 0.980 & 0.981 & 0.983 & 0.984 & 0.982 & 0.983 & 0.980 & 0.981 & 0.983 \\
\multicolumn{1}{c|}{} & \multicolumn{1}{c|}{} & \multicolumn{1}{c|}{Recall} & 0.979 & 0.979 & 0.996 & 0.992 & 0.999 & 0.998 & 0.990 & 0.992 & 0.988 & 0.990 \\
\multicolumn{1}{c|}{} & \multicolumn{1}{c|}{} & \multicolumn{1}{c|}{F1} & 0.982 & 0.979 & 0.989 & 0.987 & 0.991 & 0.990 & 0.986 & 0.986 & 0.985 & 0.986 \\
\multicolumn{1}{c|}{} & \multicolumn{1}{c|}{} & \multicolumn{1}{c|}{AUC} & 0.998 & 0.998 & 1.000 & 0.999 & 1.000 & 1.000 & 0.999 & 0.999 & 0.999 & 0.998 \\ \midrule
\multicolumn{1}{c|}{\multirow{5}{*}{\begin{tabular}[c]{@{}c@{}}Spatial \\ Depth\\ increase\end{tabular}}} & \multicolumn{1}{c|}{\multirow{5}{*}{\begin{tabular}[c]{@{}c@{}}--d\_model 512 \\ --depth\_t 2 \\ --depth\_s 4\end{tabular}}} & \multicolumn{1}{c|}{Acc} & 0.978 & 0.978 & 0.989 & 0.987 & 0.993 & 0.991 & 0.984 & 0.985 & 0.988 & 0.980 \\
\multicolumn{1}{c|}{} & \multicolumn{1}{c|}{} & \multicolumn{1}{c|}{Prec} & 0.987 & 0.986 & 0.987 & 0.986 & 0.989 & 0.987 & 0.989 & 0.982 & 0.990 & 0.983 \\
\multicolumn{1}{c|}{} & \multicolumn{1}{c|}{} & \multicolumn{1}{c|}{Recall} & 0.970 & 0.970 & 0.992 & 0.988 & 0.997 & 0.996 & 0.979 & 0.989 & 0.986 & 0.978 \\
\multicolumn{1}{c|}{} & \multicolumn{1}{c|}{} & \multicolumn{1}{c|}{F1} & 0.978 & 0.978 & 0.990 & 0.987 & 0.993 & 0.991 & 0.984 & 0.986 & 0.988 & 0.980 \\
\multicolumn{1}{c|}{} & \multicolumn{1}{c|}{} & \multicolumn{1}{c|}{AUC} & 0.998 & 0.998 & 1.000 & 0.999 & 1.000 & 1.000 & 0.999 & 0.999 & 0.999 & 0.998 \\ \midrule
\multicolumn{13}{c}{Temporal Depth} \\ \midrule
\multicolumn{1}{c|}{\multirow{5}{*}{\begin{tabular}[c]{@{}c@{}}Temporal \\ Depth \\ decrease\end{tabular}}} & \multicolumn{1}{c|}{\multirow{5}{*}{\begin{tabular}[c]{@{}c@{}}--d\_model 512 \\ --depth\_t 1 \\ --depth\_s 2\end{tabular}}} & \multicolumn{1}{c|}{Acc} & 0.981 & 0.979 & 0.988 & 0.986 & 0.991 & 0.989 & 0.987 & 0.984 & 0.988 & 0.989 \\
\multicolumn{1}{c|}{} & \multicolumn{1}{c|}{} & \multicolumn{1}{c|}{Prec} & 0.984 & 0.982 & 0.982 & 0.984 & 0.983 & 0.980 & 0.987 & 0.979 & 0.983 & 0.988 \\
\multicolumn{1}{c|}{} & \multicolumn{1}{c|}{} & \multicolumn{1}{c|}{Recall} & 0.980 & 0.977 & 0.994 & 0.988 & 0.999 & 0.998 & 0.988 & 0.990 & 0.993 & 0.990 \\
\multicolumn{1}{c|}{} & \multicolumn{1}{c|}{} & \multicolumn{1}{c|}{F1} & 0.982 & 0.979 & 0.988 & 0.986 & 0.991 & 0.989 & 0.987 & 0.984 & 0.988 & 0.989 \\
\multicolumn{1}{c|}{} & \multicolumn{1}{c|}{} & \multicolumn{1}{c|}{AUC} & 0.998 & 0.997 & 1.000 & 0.999 & 1.000 & 1.000 & 0.999 & 0.999 & 0.999 & 0.999 \\ \midrule
\multicolumn{1}{c|}{\multirow{5}{*}{\begin{tabular}[c]{@{}c@{}}Temporal \\ Depth \\ increase\end{tabular}}} & \multicolumn{1}{c|}{\multirow{5}{*}{\begin{tabular}[c]{@{}c@{}}--d\_model 512 \\ --depth\_t 4 \\ --depth\_s 2\end{tabular}}} & \multicolumn{1}{c|}{Acc} & 0.975 & 0.977 & 0.991 & 0.985 & 0.992 & 0.991 & 0.982 & 0.986 & 0.985 & 0.979 \\
\multicolumn{1}{c|}{} & \multicolumn{1}{c|}{} & \multicolumn{1}{c|}{Prec} & 0.990 & 0.987 & 0.988 & 0.988 & 0.989 & 0.988 & 0.992 & 0.986 & 0.989 & 0.989 \\
\multicolumn{1}{c|}{} & \multicolumn{1}{c|}{} & \multicolumn{1}{c|}{Recall} & 0.961 & 0.968 & 0.994 & 0.984 & 0.999 & 0.994 & 0.973 & 0.987 & 0.982 & 0.970 \\
\multicolumn{1}{c|}{} & \multicolumn{1}{c|}{} & \multicolumn{1}{c|}{F1} & 0.975 & 0.977 & 0.991 & 0.986 & 0.993 & 0.991 & 0.983 & 0.986 & 0.986 & 0.979 \\
\multicolumn{1}{c|}{} & \multicolumn{1}{c|}{} & \multicolumn{1}{c|}{AUC} & 0.998 & 0.998 & 1.000 & 0.999 & 1.000 & 1.000 & 0.999 & 0.999 & 0.999 & 0.998 \\ \midrule
\multicolumn{13}{c}{Temporal \& Spatial Depth} \\ \midrule
\multicolumn{1}{c|}{\multirow{5}{*}{\begin{tabular}[c]{@{}c@{}}T\&S \\ Depth\\ icnrease 1\end{tabular}}} & \multicolumn{1}{c|}{\multirow{5}{*}{\begin{tabular}[c]{@{}c@{}}--d\_model 512 \\ --depth\_t 3 \\ --depth\_s 3\end{tabular}}} & \multicolumn{1}{c|}{Acc} & 0.976 & 0.979 & 0.990 & 0.987 & 0.992 & 0.990 & 0.987 & 0.989 & 0.987 & 0.985 \\
\multicolumn{1}{c|}{} & \multicolumn{1}{c|}{} & \multicolumn{1}{c|}{Prec} & 0.990 & 0.988 & 0.987 & 0.990 & 0.987 & 0.988 & 0.993 & 0.988 & 0.988 & 0.989 \\
\multicolumn{1}{c|}{} & \multicolumn{1}{c|}{} & \multicolumn{1}{c|}{Recall} & 0.964 & 0.970 & 0.993 & 0.984 & 0.997 & 0.993 & 0.982 & 0.990 & 0.987 & 0.983 \\
\multicolumn{1}{c|}{} & \multicolumn{1}{c|}{} & \multicolumn{1}{c|}{F1} & 0.976 & 0.979 & 0.990 & 0.987 & 0.992 & 0.991 & 0.988 & 0.989 & 0.987 & 0.986 \\
\multicolumn{1}{c|}{} & \multicolumn{1}{c|}{} & \multicolumn{1}{c|}{AUC} & 0.998 & 0.998 & 0.999 & 0.999 & 1.000 & 1.000 & 0.999 & 0.999 & 0.999 & 0.998 \\ \midrule
\multicolumn{1}{c|}{\multirow{5}{*}{\begin{tabular}[c]{@{}c@{}}T\&S \\ Depth\\ increase 2\end{tabular}}} & \multicolumn{1}{c|}{\multirow{5}{*}{\begin{tabular}[c]{@{}c@{}}--d\_model 512 \\ --depth\_t 4 \\ --depth\_s 4\end{tabular}}} & \multicolumn{1}{c|}{Acc} & 0.975 & 0.978 & 0.989 & 0.987 & 0.991 & 0.992 & 0.985 & 0.987 & 0.991 & 0.980 \\
\multicolumn{1}{c|}{} & \multicolumn{1}{c|}{} & \multicolumn{1}{c|}{Prec} & 0.983 & 0.984 & 0.984 & 0.987 & 0.985 & 0.988 & 0.987 & 0.981 & 0.992 & 0.984 \\
\multicolumn{1}{c|}{} & \multicolumn{1}{c|}{} & \multicolumn{1}{c|}{Recall} & 0.967 & 0.972 & 0.993 & 0.987 & 0.997 & 0.996 & 0.984 & 0.993 & 0.992 & 0.976 \\
\multicolumn{1}{c|}{} & \multicolumn{1}{c|}{} & \multicolumn{1}{c|}{F1} & 0.975 & 0.978 & 0.989 & 0.987 & 0.991 & 0.992 & 0.986 & 0.987 & 0.992 & 0.980 \\
\multicolumn{1}{c|}{} & \multicolumn{1}{c|}{} & \multicolumn{1}{c|}{AUC} & 0.997 & 0.998 & 1.000 & 0.999 & 1.000 & 1.000 & 0.999 & 0.999 & 0.999 & 0.998 \\ \midrule
\multicolumn{13}{c}{Model size} \\ \midrule
\multicolumn{1}{c|}{\multirow{5}{*}{Large}} & \multicolumn{1}{c|}{\multirow{5}{*}{\begin{tabular}[c]{@{}c@{}}--d\_model 768 \\ --depth\_t 4 \\ --depth\_s 4\end{tabular}}} & \multicolumn{1}{c|}{Acc} & 0.962 & 0.949 & 0.968 & 0.975 & 0.974 & 0.973 & 0.969 & 0.963 & 0.967 & 0.963 \\
\multicolumn{1}{c|}{} & \multicolumn{1}{c|}{} & \multicolumn{1}{c|}{Prec} & 0.960 & 0.953 & 0.947 & 0.964 & 0.955 & 0.955 & 0.958 & 0.950 & 0.953 & 0.947 \\
\multicolumn{1}{c|}{} & \multicolumn{1}{c|}{} & \multicolumn{1}{c|}{Recall} & 0.965 & 0.946 & 0.993 & 0.987 & 0.997 & 0.994 & 0.983 & 0.979 & 0.984 & 0.981 \\
\multicolumn{1}{c|}{} & \multicolumn{1}{c|}{} & \multicolumn{1}{c|}{F1} & 0.963 & 0.950 & 0.969 & 0.976 & 0.976 & 0.974 & 0.970 & 0.964 & 0.968 & 0.964 \\
\multicolumn{1}{c|}{} & \multicolumn{1}{c|}{} & \multicolumn{1}{c|}{AUC} & 0.993 & 0.989 & 0.998 & 0.996 & 0.999 & 0.998 & 0.996 & 0.995 & 0.996 & 0.994 \\ \midrule
\multicolumn{1}{c|}{\multirow{5}{*}{Small}} & \multicolumn{1}{c|}{\multirow{5}{*}{\begin{tabular}[c]{@{}c@{}}--d\_model 256 \\ --depth\_t 1 \\ --depth\_s 1\end{tabular}}} & \multicolumn{1}{c|}{Acc} & 0.975 & 0.975 & 0.988 & 0.985 & 0.990 & 0.990 & 0.989 & 0.985 & 0.984 & 0.982 \\
\multicolumn{1}{c|}{} & \multicolumn{1}{c|}{} & \multicolumn{1}{c|}{Prec} & 0.987 & 0.988 & 0.984 & 0.985 & 0.984 & 0.987 & 0.992 & 0.983 & 0.992 & 0.987 \\
\multicolumn{1}{c|}{} & \multicolumn{1}{c|}{} & \multicolumn{1}{c|}{Recall} & 0.963 & 0.962 & 0.993 & 0.987 & 0.997 & 0.993 & 0.987 & 0.988 & 0.977 & 0.978 \\
\multicolumn{1}{c|}{} & \multicolumn{1}{c|}{} & \multicolumn{1}{c|}{F1} & 0.975 & 0.975 & 0.988 & 0.986 & 0.990 & 0.990 & 0.990 & 0.986 & 0.984 & 0.983 \\
\multicolumn{1}{c|}{} & \multicolumn{1}{c|}{} & \multicolumn{1}{c|}{AUC} & 0.998 & 0.998 & 0.999 & 0.999 & 1.000 & 1.000 & 0.999 & 0.999 & 0.999 & 0.998 \\ \bottomrule
\end{tabular}%
}
\end{table}

\begin{table}[]
\centering
\caption{Parameter sweep result on test set}
\label{tab:ps_test}
\resizebox{\textwidth}{!}{%
\begin{tabular}{@{}cccccccccccccccccc@{}}
\toprule
\textbf{Model} & \textbf{Configs} & \textbf{Metric} & \textbf{Unk} & \textbf{RM2} & \textbf{Kling1} & \textbf{Hailuo} & \textbf{SD} & \textbf{Mochi} & \textbf{JM} & \textbf{Gen3} & \textbf{Luma} & \textbf{Vidu} & \textbf{PRM} & \textbf{SKR} & \textbf{PV} & \textbf{Pika2} & \textbf{Gen4} \\ \midrule
\multicolumn{18}{c}{Model Dimension} \\ \midrule
\multicolumn{1}{c|}{\multirow{5}{*}{\begin{tabular}[c]{@{}c@{}}Dimension \\ reduction\end{tabular}}} & \multicolumn{1}{c|}{\multirow{5}{*}{\begin{tabular}[c]{@{}c@{}}--d\_model 256 \\ --depth\_t 2 \\ --depth\_s 2\end{tabular}}} & \multicolumn{1}{c|}{Acc} & 0.953 & 0.987 & 0.952 & 0.970 & 0.983 & 0.910 & 0.991 & 0.986 & 0.971 & 0.982 & 0.982 & 0.963 & 0.978 & 0.967 & 0.934 \\
\multicolumn{1}{c|}{} & \multicolumn{1}{c|}{} & \multicolumn{1}{c|}{Prec} & 0.981 & 0.985 & 0.985 & 0.986 & 0.982 & 0.984 & 0.986 & 0.980 & 0.992 & 0.970 & 0.980 & 0.982 & 0.975 & 0.983 & 1.000 \\
\multicolumn{1}{c|}{} & \multicolumn{1}{c|}{} & \multicolumn{1}{c|}{Recall} & 0.925 & 0.989 & 0.921 & 0.956 & 0.986 & 0.838 & 0.996 & 0.992 & 0.952 & 0.996 & 0.986 & 0.945 & 0.982 & 0.952 & 0.870 \\
\multicolumn{1}{c|}{} & \multicolumn{1}{c|}{} & \multicolumn{1}{c|}{F1} & 0.952 & 0.987 & 0.952 & 0.971 & 0.984 & 0.905 & 0.991 & 0.986 & 0.971 & 0.983 & 0.983 & 0.963 & 0.978 & 0.967 & 0.931 \\
\multicolumn{1}{c|}{} & \multicolumn{1}{c|}{} & \multicolumn{1}{c|}{AUC} & 0.991 & 0.999 & 0.993 & 0.996 & 0.998 & 0.984 & 0.998 & 0.998 & 0.998 & 0.999 & 0.999 & 0.995 & 0.998 & 0.997 & 0.997 \\ \midrule
\multicolumn{1}{c|}{\multirow{5}{*}{\begin{tabular}[c]{@{}c@{}}Dimension \\ increase\end{tabular}}} & \multicolumn{1}{c|}{\multirow{5}{*}{\begin{tabular}[c]{@{}c@{}}--d\_model 768 \\ --depth\_t 2 \\ --depth\_s 2\end{tabular}}} & \multicolumn{1}{c|}{Acc} & 0.961 & 0.987 & 0.959 & 0.980 & 0.983 & 0.920 & 0.992 & 0.985 & 0.970 & 0.979 & 0.988 & 0.976 & 0.982 & 0.975 & 0.967 \\
\multicolumn{1}{c|}{} & \multicolumn{1}{c|}{} & \multicolumn{1}{c|}{Prec} & 0.980 & 0.985 & 0.984 & 0.984 & 0.986 & 0.986 & 0.986 & 0.980 & 0.990 & 0.968 & 0.988 & 0.980 & 0.972 & 0.984 & 0.993 \\
\multicolumn{1}{c|}{} & \multicolumn{1}{c|}{} & \multicolumn{1}{c|}{Recall} & 0.943 & 0.989 & 0.936 & 0.977 & 0.981 & 0.855 & 0.998 & 0.990 & 0.952 & 0.992 & 0.990 & 0.972 & 0.993 & 0.968 & 0.942 \\
\multicolumn{1}{c|}{} & \multicolumn{1}{c|}{} & \multicolumn{1}{c|}{F1} & 0.961 & 0.987 & 0.959 & 0.980 & 0.983 & 0.916 & 0.992 & 0.985 & 0.970 & 0.980 & 0.989 & 0.976 & 0.982 & 0.976 & 0.967 \\
\multicolumn{1}{c|}{} & \multicolumn{1}{c|}{} & \multicolumn{1}{c|}{AUC} & 0.990 & 0.999 & 0.995 & 0.997 & 0.999 & 0.990 & 0.999 & 0.999 & 0.996 & 0.999 & 0.999 & 0.995 & 0.999 & 0.999 & 0.998 \\ \midrule
\multicolumn{18}{c}{Spatial Depth} \\ \midrule
\multicolumn{1}{c|}{\multirow{5}{*}{\begin{tabular}[c]{@{}c@{}}Spatial \\ Depth\\ decrease\end{tabular}}} & \multicolumn{1}{c|}{\multirow{5}{*}{\begin{tabular}[c]{@{}c@{}}--d\_model 512 \\ --depth\_t 2 \\ --depth\_s 1\end{tabular}}} & \multicolumn{1}{c|}{Acc} & 0.954 & 0.989 & 0.960 & 0.977 & 0.976 & 0.918 & 0.987 & 0.987 & 0.978 & 0.980 & 0.990 & 0.974 & 0.978 & 0.973 & 0.970 \\
\multicolumn{1}{c|}{} & \multicolumn{1}{c|}{} & \multicolumn{1}{c|}{Prec} & 0.980 & 0.985 & 0.983 & 0.986 & 0.980 & 0.984 & 0.978 & 0.980 & 0.984 & 0.970 & 0.986 & 0.982 & 0.968 & 0.984 & 0.993 \\
\multicolumn{1}{c|}{} & \multicolumn{1}{c|}{} & \multicolumn{1}{c|}{Recall} & 0.928 & 0.993 & 0.939 & 0.968 & 0.973 & 0.853 & 0.996 & 0.994 & 0.972 & 0.992 & 0.994 & 0.967 & 0.989 & 0.962 & 0.948 \\
\multicolumn{1}{c|}{} & \multicolumn{1}{c|}{} & \multicolumn{1}{c|}{F1} & 0.954 & 0.989 & 0.960 & 0.977 & 0.977 & 0.914 & 0.987 & 0.987 & 0.978 & 0.981 & 0.990 & 0.975 & 0.979 & 0.973 & 0.970 \\
\multicolumn{1}{c|}{} & \multicolumn{1}{c|}{} & \multicolumn{1}{c|}{AUC} & 0.990 & 0.999 & 0.993 & 0.997 & 0.998 & 0.988 & 0.999 & 0.998 & 0.998 & 0.999 & 1.000 & 0.997 & 0.999 & 0.999 & 0.999 \\ \midrule
\multicolumn{1}{c|}{\multirow{5}{*}{\begin{tabular}[c]{@{}c@{}}Spatial \\ Depth\\ increase\end{tabular}}} & \multicolumn{1}{c|}{\multirow{5}{*}{\begin{tabular}[c]{@{}c@{}}--d\_model 512 \\ --depth\_t 2 \\ --depth\_s 4\end{tabular}}} & \multicolumn{1}{c|}{Acc} & 0.951 & 0.988 & 0.954 & 0.977 & 0.983 & 0.902 & 0.993 & 0.985 & 0.963 & 0.987 & 0.990 & 0.965 & 0.987 & 0.962 & 0.947 \\
\multicolumn{1}{c|}{} & \multicolumn{1}{c|}{} & \multicolumn{1}{c|}{Prec} & 0.987 & 0.989 & 0.989 & 0.991 & 0.988 & 0.986 & 0.990 & 0.986 & 0.987 & 0.980 & 0.990 & 0.984 & 0.989 & 0.989 & 1.000 \\
\multicolumn{1}{c|}{} & \multicolumn{1}{c|}{} & \multicolumn{1}{c|}{Recall} & 0.916 & 0.988 & 0.920 & 0.964 & 0.980 & 0.821 & 0.996 & 0.984 & 0.940 & 0.996 & 0.990 & 0.947 & 0.986 & 0.935 & 0.896 \\
\multicolumn{1}{c|}{} & \multicolumn{1}{c|}{} & \multicolumn{1}{c|}{F1} & 0.950 & 0.989 & 0.954 & 0.978 & 0.984 & 0.896 & 0.993 & 0.985 & 0.963 & 0.988 & 0.990 & 0.966 & 0.987 & 0.961 & 0.945 \\
\multicolumn{1}{c|}{} & \multicolumn{1}{c|}{} & \multicolumn{1}{c|}{AUC} & 0.989 & 0.999 & 0.994 & 0.997 & 0.998 & 0.988 & 1.000 & 0.997 & 0.997 & 0.999 & 0.999 & 0.997 & 0.998 & 0.998 & 0.999 \\ \midrule
\multicolumn{18}{c}{Temporal Depth} \\ \midrule
\multicolumn{1}{c|}{\multirow{5}{*}{\begin{tabular}[c]{@{}c@{}}Temporal \\ Depth \\ decrease\end{tabular}}} & \multicolumn{1}{c|}{\multirow{5}{*}{\begin{tabular}[c]{@{}c@{}}--d\_model 512 \\ --depth\_t 1 \\ --depth\_s 2\end{tabular}}} & \multicolumn{1}{c|}{Acc} & 0.964 & 0.984 & 0.959 & 0.982 & 0.985 & 0.925 & 0.993 & 0.985 & 0.974 & 0.982 & 0.987 & 0.972 & 0.982 & 0.962 & 0.947 \\
\multicolumn{1}{c|}{} & \multicolumn{1}{c|}{} & \multicolumn{1}{c|}{Prec} & 0.983 & 0.983 & 0.987 & 0.987 & 0.984 & 0.984 & 0.988 & 0.976 & 0.992 & 0.967 & 0.986 & 0.982 & 0.975 & 0.978 & 1.000 \\
\multicolumn{1}{c|}{} & \multicolumn{1}{c|}{} & \multicolumn{1}{c|}{Recall} & 0.946 & 0.987 & 0.932 & 0.978 & 0.986 & 0.867 & 0.998 & 0.994 & 0.958 & 1.000 & 0.990 & 0.962 & 0.989 & 0.946 & 0.896 \\
\multicolumn{1}{c|}{} & \multicolumn{1}{c|}{} & \multicolumn{1}{c|}{F1} & 0.964 & 0.985 & 0.959 & 0.983 & 0.985 & 0.922 & 0.993 & 0.985 & 0.975 & 0.983 & 0.988 & 0.972 & 0.982 & 0.962 & 0.945 \\
\multicolumn{1}{c|}{} & \multicolumn{1}{c|}{} & \multicolumn{1}{c|}{AUC} & 0.992 & 0.999 & 0.992 & 0.997 & 0.999 & 0.989 & 0.999 & 0.998 & 0.998 & 0.999 & 0.999 & 0.996 & 0.999 & 0.997 & 0.999 \\ \midrule
\multicolumn{1}{c|}{\multirow{5}{*}{\begin{tabular}[c]{@{}c@{}}Temporal \\ Depth \\ increase\end{tabular}}} & \multicolumn{1}{c|}{\multirow{5}{*}{\begin{tabular}[c]{@{}c@{}}--d\_model 512 \\ --depth\_t 4 \\ --depth\_s 2\end{tabular}}} & \multicolumn{1}{c|}{Acc} & 0.951 & 0.985 & 0.948 & 0.975 & 0.983 & 0.887 & 0.993 & 0.985 & 0.963 & 0.980 & 0.986 & 0.956 & 0.983 & 0.975 & 0.924 \\
\multicolumn{1}{c|}{} & \multicolumn{1}{c|}{} & \multicolumn{1}{c|}{Prec} & 0.989 & 0.989 & 0.992 & 0.992 & 0.992 & 0.991 & 0.992 & 0.988 & 0.996 & 0.978 & 0.992 & 0.987 & 0.986 & 0.994 & 1.000 \\
\multicolumn{1}{c|}{} & \multicolumn{1}{c|}{} & \multicolumn{1}{c|}{Recall} & 0.915 & 0.981 & 0.905 & 0.959 & 0.975 & 0.786 & 0.994 & 0.982 & 0.932 & 0.984 & 0.982 & 0.927 & 0.982 & 0.957 & 0.851 \\
\multicolumn{1}{c|}{} & \multicolumn{1}{c|}{} & \multicolumn{1}{c|}{F1} & 0.951 & 0.985 & 0.947 & 0.975 & 0.984 & 0.877 & 0.993 & 0.985 & 0.963 & 0.981 & 0.987 & 0.956 & 0.984 & 0.975 & 0.919 \\
\multicolumn{1}{c|}{} & \multicolumn{1}{c|}{} & \multicolumn{1}{c|}{AUC} & 0.992 & 0.999 & 0.993 & 0.997 & 0.998 & 0.987 & 0.999 & 0.999 & 0.996 & 0.999 & 0.999 & 0.995 & 0.999 & 0.999 & 0.996 \\ \midrule
\multicolumn{18}{c}{Temporal \& Spatial Depth} \\ \midrule
\multicolumn{1}{c|}{\multirow{5}{*}{\begin{tabular}[c]{@{}c@{}}T\&S \\ Depth\\ icnrease 1\end{tabular}}} & \multicolumn{1}{c|}{\multirow{5}{*}{\begin{tabular}[c]{@{}c@{}}--d\_model 512 \\ --depth\_t 3 \\ --depth\_s 3\end{tabular}}} & \multicolumn{1}{c|}{Acc} & 0.947 & 0.987 & 0.950 & 0.971 & 0.980 & 0.898 & 0.991 & 0.987 & 0.961 & 0.983 & 0.986 & 0.958 & 0.980 & 0.967 & 0.940 \\
\multicolumn{1}{c|}{} & \multicolumn{1}{c|}{} & \multicolumn{1}{c|}{Prec} & 0.988 & 0.987 & 0.990 & 0.993 & 0.987 & 0.994 & 0.990 & 0.988 & 0.989 & 0.976 & 0.994 & 0.987 & 0.985 & 0.994 & 1.000 \\
\multicolumn{1}{c|}{} & \multicolumn{1}{c|}{} & \multicolumn{1}{c|}{Recall} & 0.906 & 0.987 & 0.910 & 0.949 & 0.974 & 0.805 & 0.992 & 0.986 & 0.934 & 0.992 & 0.980 & 0.930 & 0.975 & 0.941 & 0.883 \\
\multicolumn{1}{c|}{} & \multicolumn{1}{c|}{} & \multicolumn{1}{c|}{F1} & 0.945 & 0.987 & 0.949 & 0.971 & 0.980 & 0.890 & 0.991 & 0.987 & 0.961 & 0.984 & 0.987 & 0.957 & 0.980 & 0.967 & 0.938 \\
\multicolumn{1}{c|}{} & \multicolumn{1}{c|}{} & \multicolumn{1}{c|}{AUC} & 0.990 & 0.999 & 0.993 & 0.997 & 0.998 & 0.986 & 0.999 & 0.999 & 0.997 & 0.999 & 0.998 & 0.996 & 0.998 & 0.997 & 0.998 \\ \midrule
\multicolumn{1}{c|}{\multirow{5}{*}{\begin{tabular}[c]{@{}c@{}}T\&S \\ Depth\\ increase 2\end{tabular}}} & \multicolumn{1}{c|}{\multirow{5}{*}{\begin{tabular}[c]{@{}c@{}}--d\_model 512 \\ --depth\_t 4 \\ --depth\_s 4\end{tabular}}} & \multicolumn{1}{c|}{Acc} & 0.955 & 0.984 & 0.953 & 0.981 & 0.990 & 0.889 & 0.988 & 0.986 & 0.963 & 0.983 & 0.986 & 0.967 & 0.976 & 0.975 & 0.944 \\
\multicolumn{1}{c|}{} & \multicolumn{1}{c|}{} & \multicolumn{1}{c|}{Prec} & 0.986 & 0.986 & 0.986 & 0.989 & 0.991 & 0.985 & 0.986 & 0.984 & 0.989 & 0.976 & 0.988 & 0.987 & 0.978 & 0.994 & 1.000 \\
\multicolumn{1}{c|}{} & \multicolumn{1}{c|}{} & \multicolumn{1}{c|}{Recall} & 0.925 & 0.982 & 0.920 & 0.973 & 0.988 & 0.795 & 0.990 & 0.988 & 0.938 & 0.992 & 0.986 & 0.947 & 0.975 & 0.957 & 0.890 \\
\multicolumn{1}{c|}{} & \multicolumn{1}{c|}{} & \multicolumn{1}{c|}{F1} & 0.954 & 0.984 & 0.952 & 0.981 & 0.990 & 0.880 & 0.988 & 0.986 & 0.963 & 0.984 & 0.987 & 0.967 & 0.976 & 0.975 & 0.942 \\
\multicolumn{1}{c|}{} & \multicolumn{1}{c|}{} & \multicolumn{1}{c|}{AUC} & 0.991 & 0.999 & 0.992 & 0.997 & 0.999 & 0.982 & 0.999 & 0.998 & 0.996 & 0.999 & 0.999 & 0.996 & 0.997 & 0.999 & 0.999 \\ \midrule
\multicolumn{18}{c}{Model size} \\ \midrule
\multicolumn{1}{c|}{\multirow{5}{*}{Large}} & \multicolumn{1}{c|}{\multirow{5}{*}{\begin{tabular}[c]{@{}c@{}}--d\_model 768 \\ --depth\_t 4 \\ --depth\_s 4\end{tabular}}} & \multicolumn{1}{c|}{Acc} & 0.938 & 0.976 & 0.945 & 0.966 & 0.970 & 0.885 & 0.965 & 0.969 & 0.943 & 0.976 & 0.976 & 0.889 & 0.937 & 0.918 & 0.827 \\
\multicolumn{1}{c|}{} & \multicolumn{1}{c|}{} & \multicolumn{1}{c|}{Prec} & 0.955 & 0.960 & 0.958 & 0.965 & 0.963 & 0.948 & 0.936 & 0.961 & 0.951 & 0.959 & 0.960 & 0.959 & 0.939 & 0.948 & 0.955 \\
\multicolumn{1}{c|}{} & \multicolumn{1}{c|}{} & \multicolumn{1}{c|}{Recall} & 0.923 & 0.993 & 0.932 & 0.967 & 0.978 & 0.821 & 1.000 & 0.980 & 0.936 & 0.996 & 0.994 & 0.817 & 0.939 & 0.887 & 0.695 \\
\multicolumn{1}{c|}{} & \multicolumn{1}{c|}{} & \multicolumn{1}{c|}{F1} & 0.939 & 0.977 & 0.945 & 0.966 & 0.971 & 0.880 & 0.967 & 0.970 & 0.944 & 0.977 & 0.977 & 0.882 & 0.939 & 0.917 & 0.805 \\
\multicolumn{1}{c|}{} & \multicolumn{1}{c|}{} & \multicolumn{1}{c|}{AUC} & 0.983 & 0.997 & 0.985 & 0.993 & 0.995 & 0.960 & 0.999 & 0.996 & 0.990 & 0.999 & 0.998 & 0.967 & 0.986 & 0.980 & 0.939 \\ \midrule
\multicolumn{1}{c|}{\multirow{5}{*}{Small}} & \multicolumn{1}{c|}{\multirow{5}{*}{\begin{tabular}[c]{@{}c@{}}--d\_model 256 \\ --depth\_t 1 \\ --depth\_s 1\end{tabular}}} & \multicolumn{1}{c|}{Acc} & 0.944 & 0.986 & 0.941 & 0.961 & 0.977 & 0.892 & 0.987 & 0.980 & 0.961 & 0.981 & 0.984 & 0.967 & 0.976 & 0.959 & 0.934 \\
\multicolumn{1}{c|}{} & \multicolumn{1}{c|}{} & \multicolumn{1}{c|}{Prec} & 0.987 & 0.988 & 0.991 & 0.989 & 0.989 & 0.987 & 0.984 & 0.984 & 0.989 & 0.978 & 0.986 & 0.992 & 0.978 & 0.989 & 1.000 \\
\multicolumn{1}{c|}{} & \multicolumn{1}{c|}{} & \multicolumn{1}{c|}{Recall} & 0.902 & 0.983 & 0.892 & 0.934 & 0.967 & 0.800 & 0.990 & 0.976 & 0.934 & 0.986 & 0.984 & 0.942 & 0.975 & 0.930 & 0.870 \\
\multicolumn{1}{c|}{} & \multicolumn{1}{c|}{} & \multicolumn{1}{c|}{F1} & 0.942 & 0.986 & 0.939 & 0.960 & 0.978 & 0.884 & 0.987 & 0.980 & 0.961 & 0.982 & 0.985 & 0.967 & 0.976 & 0.958 & 0.931 \\
\multicolumn{1}{c|}{} & \multicolumn{1}{c|}{} & \multicolumn{1}{c|}{AUC} & 0.988 & 0.999 & 0.992 & 0.996 & 0.998 & 0.983 & 0.999 & 0.998 & 0.996 & 0.999 & 0.999 & 0.995 & 0.998 & 0.997 & 0.998 \\ \bottomrule
\end{tabular}%
}
\end{table}

\begin{table}[]
\centering
\caption{Ablation of VJEPA-2 val set}
\label{tab:vjp_val}
\resizebox{\textwidth}{!}{%
\begin{tabular}{@{}clllllllllll|l@{}}
\toprule
\multicolumn{1}{l}{\textbf{Model}} & \textbf{Metric} & \multicolumn{1}{c}{\textbf{Veo3}} & \multicolumn{1}{c}{\textbf{Sora2}} & \multicolumn{1}{c}{\textbf{HY}} & \multicolumn{1}{c}{\textbf{CVX}} & \multicolumn{1}{c}{\textbf{EA}} & \multicolumn{1}{c}{\textbf{LTX}} & \multicolumn{1}{c}{\textbf{Pika1}} & \multicolumn{1}{c}{\textbf{Wan2}} & \multicolumn{1}{c}{\textbf{Kling2}} & \multicolumn{1}{c|}{\textbf{Sora}} & \textbf{Avg} \\ \midrule
\multicolumn{1}{c|}{\multirow{5}{*}{\textbf{Base}}} & \multicolumn{1}{l|}{Acc} & \multicolumn{1}{r}{0.984} & \multicolumn{1}{r}{0.982} & \multicolumn{1}{r}{0.989} & \multicolumn{1}{r}{0.986} & \multicolumn{1}{r}{0.991} & \multicolumn{1}{r}{\textbf{0.987}} & \multicolumn{1}{r}{0.989} & \multicolumn{1}{r}{0.985} & \multicolumn{1}{r}{0.988} & \multicolumn{1}{r|}{0.985} & \multicolumn{1}{r}{0.999} \\
\multicolumn{1}{c|}{} & \multicolumn{1}{l|}{Prec} & \multicolumn{1}{r}{0.985} & \multicolumn{1}{r}{0.980} & \multicolumn{1}{r}{0.981} & \multicolumn{1}{r}{0.981} & \multicolumn{1}{r}{0.983} & \multicolumn{1}{r}{0.978} & \multicolumn{1}{r}{0.989} & \multicolumn{1}{r}{0.978} & \multicolumn{1}{r}{0.987} & \multicolumn{1}{r|}{0.985} & \multicolumn{1}{r}{0.987} \\
\multicolumn{1}{c|}{} & \multicolumn{1}{l|}{Recall} & \multicolumn{1}{r}{0.982} & \multicolumn{1}{r}{0.984} & \multicolumn{1}{r}{0.997} & \multicolumn{1}{r}{0.992} & \multicolumn{1}{r}{0.999} & \multicolumn{1}{r}{0.997} & \multicolumn{1}{r}{0.990} & \multicolumn{1}{r}{0.993} & \multicolumn{1}{r}{0.990} & \multicolumn{1}{r|}{0.986} & \multicolumn{1}{r}{0.983} \\
\multicolumn{1}{c|}{} & \multicolumn{1}{l|}{F1} & \multicolumn{1}{r}{0.984} & \multicolumn{1}{r}{0.982} & \multicolumn{1}{r}{0.989} & \multicolumn{1}{r}{0.987} & \multicolumn{1}{r}{0.991} & \multicolumn{1}{r}{0.988} & \multicolumn{1}{r}{0.989} & \multicolumn{1}{r}{0.986} & \multicolumn{1}{r}{0.989} & \multicolumn{1}{r|}{0.986} & \multicolumn{1}{r}{0.991} \\
\bottomrule
\end{tabular}%
}
\end{table}

\begin{table}[]
\centering
\caption{Ablation of VJEPA-2 test set}
\label{tab:vjp_test}
\resizebox{\textwidth}{!}{%
\begin{tabular}{@{}cllllllllllllllll|l@{}}
\toprule
\multicolumn{1}{l}{\textbf{Model}} & \textbf{Metric} & \multicolumn{1}{c}{\textbf{Unk}} & \multicolumn{1}{c}{\textbf{RM2}} & \multicolumn{1}{c}{\textbf{Kling1}} & \multicolumn{1}{c}{\textbf{Hailuo}} & \multicolumn{1}{c}{\textbf{SD}} & \multicolumn{1}{c}{\textbf{Mochi}} & \multicolumn{1}{c}{\textbf{JM}} & \multicolumn{1}{c}{\textbf{Gen3}} & \multicolumn{1}{c}{\textbf{Luma}} & \multicolumn{1}{c}{\textbf{Vidu}} & \textbf{PRM} & \textbf{SKR} & \textbf{PV} & \textbf{Pika2} & \textbf{Gen4} & \textbf{Avg} \\ \midrule
\multicolumn{1}{c|}{\multirow{5}{*}{\textbf{Base}}} & \multicolumn{1}{l|}{Acc} & \multicolumn{1}{r}{0.961} & \multicolumn{1}{r}{0.987} & \multicolumn{1}{r}{0.959} & \multicolumn{1}{r}{0.980} & \multicolumn{1}{r}{0.983} & \multicolumn{1}{r}{0.920} & \multicolumn{1}{r}{0.992} & \multicolumn{1}{r}{0.985} & \multicolumn{1}{r}{0.970} & \multicolumn{1}{r}{0.979} & \multicolumn{1}{r}{0.988} & \multicolumn{1}{r}{0.976} & \multicolumn{1}{r}{0.982} & \multicolumn{1}{r}{0.975} & \multicolumn{1}{r|}{0.967} & \multicolumn{1}{r}{0.974} \\
\multicolumn{1}{c|}{} & \multicolumn{1}{l|}{Prec} & \multicolumn{1}{r}{0.980} & \multicolumn{1}{r}{0.985} & \multicolumn{1}{r}{0.984} & \multicolumn{1}{r}{0.984} & \multicolumn{1}{r}{0.986} & \multicolumn{1}{r}{0.986} & \multicolumn{1}{r}{0.986} & \multicolumn{1}{r}{0.980} & \multicolumn{1}{r}{0.990} & \multicolumn{1}{r}{0.968} & \multicolumn{1}{r}{0.988} & \multicolumn{1}{r}{0.980} & \multicolumn{1}{r}{0.972} & \multicolumn{1}{r}{0.984} & \multicolumn{1}{r|}{0.993} & \multicolumn{1}{r}{0.983} \\
\bottomrule
\end{tabular}%
}
\end{table}

\begin{table}[]
\centering
\caption{Ablation of CLIP val set}
\label{tab:clip_val}
\resizebox{\textwidth}{!}{%
\begin{tabular}{@{}cllllllllllll@{}}
\toprule
\multicolumn{1}{l}{\textbf{Model}} & \textbf{Metric} & \multicolumn{1}{c}{\textbf{Veo3}} & \multicolumn{1}{c}{\textbf{Sora2}} & \multicolumn{1}{c}{\textbf{HY}} & \multicolumn{1}{c}{\textbf{CVX}} & \multicolumn{1}{c}{\textbf{EA}} & \multicolumn{1}{c}{\textbf{LTX}} & \multicolumn{1}{c}{\textbf{Pika1}} & \multicolumn{1}{c}{\textbf{Wan2}} & \multicolumn{1}{c}{\textbf{Kling2}} & \multicolumn{1}{c}{\textbf{Sora}} & \textbf{Avg} \\ \midrule
\multicolumn{1}{c|}{\multirow{5}{*}{\textbf{Base}}} & \multicolumn{1}{l|}{Acc} & 0.984 & 0.982 & 0.989 & 0.986 & 0.991 & 0.987 & 0.989 & 0.985 & 0.988 & \multicolumn{1}{l|}{0.985} & 0.971 \\
\multicolumn{1}{c|}{} & \multicolumn{1}{l|}{Prec} & 0.985 & 0.980 & 0.981 & 0.981 & 0.983 & 0.978 & 0.989 & 0.978 & 0.987 & \multicolumn{1}{l|}{0.985} & 0.976 \\
\multicolumn{1}{c|}{} & \multicolumn{1}{l|}{Recall} & 0.982 & 0.984 & 0.997 & 0.992 & 0.999 & 0.997 & 0.990 & 0.993 & 0.990 & \multicolumn{1}{l|}{0.986} & 0.968 \\
\multicolumn{1}{c|}{} & \multicolumn{1}{l|}{F1} & 0.984 & 0.982 & 0.989 & 0.987 & 0.991 & 0.988 & 0.989 & 0.986 & 0.989 & \multicolumn{1}{l|}{0.986} & 0.972 \\
\multicolumn{1}{c|}{} & \multicolumn{1}{l|}{AUC} & 0.998 & 0.998 & 1.000 & 0.999 & 1.000 & 1.000 & 0.999 & 0.999 & 0.999 & \multicolumn{1}{l|}{0.997} & 0.995 \\ \midrule
\multicolumn{1}{c|}{\multirow{5}{*}{\textbf{Ablation 1}}} & \multicolumn{1}{l|}{Acc} & 0.988 & 0.976 & 0.991 & 0.991 & 0.993 & 0.990 & 0.988 & 0.978 & 0.990 & \multicolumn{1}{l|}{0.981} & 0.970 \\
\multicolumn{1}{c|}{} & \multicolumn{1}{l|}{Prec} & 0.986 & 0.987 & 0.989 & 0.989 & 0.986 & 0.987 & 0.986 & 0.994 & 0.990 & \multicolumn{1}{l|}{0.987} & 0.973 \\
\bottomrule
\end{tabular}%
}
\end{table}

\begin{table}[]
\centering
\caption{Ablation of CLIP val set}
\label{tab:clip_val}
\resizebox{\textwidth}{!}{%
\begin{tabular}{@{}cllllllllllllllll|r@{}}
\toprule
\multicolumn{1}{l}{\textbf{Model}} & \textbf{Metric} & \multicolumn{1}{c}{\textbf{Unk}} & \multicolumn{1}{c}{\textbf{RM2}} & \multicolumn{1}{c}{\textbf{Kling1}} & \multicolumn{1}{c}{\textbf{Hailuo}} & \multicolumn{1}{c}{\textbf{SD}} & \multicolumn{1}{c}{\textbf{Mochi}} & \multicolumn{1}{c}{\textbf{JM}} & \multicolumn{1}{c}{\textbf{Gen3}} & \multicolumn{1}{c}{\textbf{Luma}} & \multicolumn{1}{c}{\textbf{Vidu}} & \textbf{PRM} & \textbf{SKR} & \textbf{PV} & \textbf{Pika2} & \textbf{Gen4} & \multicolumn{1}{l}{\textbf{Avg}} \\ \midrule
\multicolumn{1}{c|}{\multirow{5}{*}{\textbf{Base}}} & \multicolumn{1}{l|}{Acc} & 0.961 & 0.987 & 0.959 & 0.980 & 0.983 & 0.920 & 0.992 & 0.985 & 0.970 & 0.979 & 0.988 & 0.976 & 0.982 & 0.975 & 0.967 & 0.974 \\
\multicolumn{1}{c|}{} & \multicolumn{1}{l|}{Prec} & 0.980 & 0.985 & 0.984 & 0.984 & 0.986 & 0.986 & 0.986 & 0.980 & 0.990 & 0.968 & 0.988 & 0.980 & 0.972 & 0.984 & 0.993 & 0.983 \\
\multicolumn{1}{c|}{} & \multicolumn{1}{l|}{Recall} & 0.943 & 0.989 & 0.936 & 0.977 & 0.981 & 0.855 & 0.998 & 0.990 & 0.952 & 0.992 & 0.990 & 0.972 & 0.993 & 0.968 & 0.942 & 0.965 \\
\multicolumn{1}{c|}{} & \multicolumn{1}{l|}{F1} & 0.961 & 0.987 & 0.959 & 0.980 & 0.983 & 0.916 & 0.992 & 0.985 & 0.970 & 0.980 & 0.989 & 0.976 & 0.982 & 0.976 & 0.967 & 0.974 \\
\multicolumn{1}{c|}{} & \multicolumn{1}{l|}{AUC} & 0.990 & 0.999 & 0.995 & 0.997 & 0.999 & 0.990 & 0.999 & 0.999 & 0.996 & 0.999 & 0.999 & 0.995 & 0.999 & 0.999 & 0.998 & 0.997 \\ \midrule
\multicolumn{1}{c|}{\multirow{5}{*}{\textbf{Ablation 1}}} & \multicolumn{1}{l|}{Acc} & 0.952 & 0.994 & 0.965 & 0.981 & 0.985 & \textbf{0.783} & 0.993 & 0.989 & 0.982 & 0.995 & 0.992 & \textbf{0.716} & 0.924 & \textbf{0.871} & \textbf{0.721} & 0.923 \\
\multicolumn{1}{c|}{} & \multicolumn{1}{l|}{Prec} & 0.979 & 0.990 & 0.994 & 0.978 & 0.985 & 0.977 & 0.986 & 0.980 & 0.982 & 0.996 & 0.990 & 0.963 & 0.980 & 0.986 & 0.961 & 0.982 \\
\bottomrule
\end{tabular}%
}
\end{table}

\begin{table}[]
\centering
\caption{Ablation of DINO-v3 val set}
\label{tab:dino3_val}
\resizebox{\textwidth}{!}{%
\begin{tabular}{@{}cllllllllllll@{}}
\toprule
\multicolumn{1}{l}{\textbf{Model}} & \textbf{Metric} & \multicolumn{1}{c}{\textbf{Veo3}} & \multicolumn{1}{c}{\textbf{Sora2}} & \multicolumn{1}{c}{\textbf{HY}} & \multicolumn{1}{c}{\textbf{CVX}} & \multicolumn{1}{c}{\textbf{EA}} & \multicolumn{1}{c}{\textbf{LTX}} & \multicolumn{1}{c}{\textbf{Pika1}} & \multicolumn{1}{c}{\textbf{Wan2}} & \multicolumn{1}{c}{\textbf{Kling2}} & \multicolumn{1}{c}{\textbf{Sora}} & \textbf{Avg} \\ \midrule
\multicolumn{1}{c|}{\multirow{5}{*}{\textbf{Base}}} & \multicolumn{1}{l|}{Acc} & 0.963 & 0.960 & 0.978 & 0.972 & 0.989 & 0.983 & 0.978 & 0.957 & 0.988 & \multicolumn{1}{l|}{0.945} & 0.971 \\
\multicolumn{1}{c|}{} & \multicolumn{1}{l|}{Prec} & 0.979 & 0.973 & 0.977 & 0.975 & 0.980 & 0.973 & 0.977 & 0.975 & 0.981 & \multicolumn{1}{l|}{0.972} & 0.976 \\
\multicolumn{1}{c|}{} & \multicolumn{1}{l|}{Recall} & 0.947 & 0.948 & 0.980 & 0.971 & 0.999 & 0.994 & 0.980 & 0.941 & 0.996 & \multicolumn{1}{l|}{0.919} & 0.968 \\
\multicolumn{1}{c|}{} & \multicolumn{1}{l|}{F1} & 0.963 & 0.960 & 0.978 & 0.973 & 0.990 & 0.983 & 0.978 & 0.958 & 0.989 & \multicolumn{1}{l|}{0.945} & 0.972 \\
\multicolumn{1}{c|}{} & \multicolumn{1}{l|}{AUC} & 0.994 & 0.991 & 0.997 & 0.994 & 1.000 & 0.999 & 0.997 & 0.987 & 1.000 & \multicolumn{1}{l|}{0.986} & 0.995 \\ \midrule
\multicolumn{1}{c|}{\multirow{5}{*}{\textbf{Ablation 1}}} & \multicolumn{1}{l|}{Acc} & 0.960 & 0.959 & 0.978 & 0.972 & 0.989 & 0.980 & 0.974 & 0.956 & 0.986 & \multicolumn{1}{l|}{0.948} & 0.970 \\
\multicolumn{1}{c|}{} & \multicolumn{1}{l|}{Prec} & 0.975 & 0.971 & 0.974 & 0.973 & 0.979 & 0.967 & 0.972 & 0.972 & 0.978 & \multicolumn{1}{l|}{0.974} & 0.973 \\
\bottomrule
\end{tabular}%
}
\end{table}

\begin{table}[]
\centering
\caption{Ablation of DINO-v3 test set}
\label{tab:dino3_test}
\resizebox{\textwidth}{!}{%
\begin{tabular}{@{}cllllllllllllllll|l@{}}
\toprule
\multicolumn{1}{l}{\textbf{Model}} & \textbf{Metric} & \multicolumn{1}{c}{\textbf{Unk}} & \multicolumn{1}{c}{\textbf{RM2}} & \multicolumn{1}{c}{\textbf{Kling1}} & \multicolumn{1}{c}{\textbf{Hailuo}} & \multicolumn{1}{c}{\textbf{SD}} & \multicolumn{1}{c}{\textbf{Mochi}} & \multicolumn{1}{c}{\textbf{JM}} & \multicolumn{1}{c}{\textbf{Gen3}} & \multicolumn{1}{c}{\textbf{Luma}} & \multicolumn{1}{c}{\textbf{Vidu}} & \textbf{PRM} & \textbf{SKR} & \textbf{PV} & \textbf{Pika2} & \textbf{Gen4} & \textbf{Avg} \\ \midrule
\multicolumn{1}{c|}{\multirow{5}{*}{\textbf{Base}}} & \multicolumn{1}{l|}{Acc} & 0.931 & 0.990 & 0.848 & 0.976 & 0.973 & \textbf{0.770} & 0.974 & 0.962 & 0.969 & 0.980 & 0.969 & 0.683 & 0.915 & \textbf{0.794} & \textbf{0.638} & 0.891 \\
\multicolumn{1}{c|}{} & \multicolumn{1}{l|}{Prec} & 0.973 & 0.982 & 0.977 & 0.984 & 0.971 & 0.957 & 0.972 & 0.977 & 0.978 & 0.984 & 0.960 & 0.942 & 0.971 & 0.937 & 0.979 & 0.970 \\
\multicolumn{1}{c|}{} & \multicolumn{1}{l|}{Recall} & 0.889 & 0.998 & 0.719 & 0.968 & 0.976 & 0.576 & 0.976 & 0.948 & 0.962 & 0.978 & 0.980 & 0.404 & 0.859 & 0.640 & 0.299 & 0.811 \\
\multicolumn{1}{c|}{} & \multicolumn{1}{l|}{F1} & 0.929 & 0.990 & 0.828 & 0.976 & 0.974 & 0.719 & 0.974 & 0.962 & 0.970 & 0.981 & 0.970 & 0.565 & 0.912 & 0.760 & 0.458 & 0.865 \\
\multicolumn{1}{c|}{} & \multicolumn{1}{l|}{AUC} & 0.981 & 1.000 & 0.960 & 0.997 & 0.996 & 0.906 & 0.996 & 0.994 & 0.996 & 0.997 & 0.996 & 0.909 & 0.979 & 0.929 & 0.912 & 0.970 \\ \midrule
\multicolumn{1}{c|}{\multirow{5}{*}{\textbf{Ablation 1}}} & \multicolumn{1}{l|}{Acc} & 0.933 & 0.988 & 0.849 & 0.971 & 0.973 & \textbf{0.766} & 0.974 & 0.958 & 0.971 & 0.981 & 0.973 & \textbf{0.693} & 0.923 & \textbf{0.780} & \textbf{0.665} & 0.893 \\
\multicolumn{1}{c|}{} & \multicolumn{1}{l|}{Prec} & 0.972 & 0.979 & 0.977 & 0.977 & 0.969 & 0.944 & 0.968 & 0.969 & 0.976 & 0.986 & 0.962 & 0.939 & 0.976 & 0.914 & 0.982 & 0.966 \\
\bottomrule
\end{tabular}%
}
\end{table}

\begin{table}[]
\centering
\caption{Ablation of DINO-v2 val set}
\label{tab:dino2_val}
\resizebox{\textwidth}{!}{%
\begin{tabular}{@{}cllllllllllll@{}}
\toprule
\multicolumn{1}{l}{\textbf{Model}} & \textbf{Metric} & \multicolumn{1}{c}{\textbf{Veo3}} & \multicolumn{1}{c}{\textbf{Sora2}} & \multicolumn{1}{c}{\textbf{HY}} & \multicolumn{1}{c}{\textbf{CVX}} & \multicolumn{1}{c}{\textbf{EA}} & \multicolumn{1}{c}{\textbf{LTX}} & \multicolumn{1}{c}{\textbf{Pika1}} & \multicolumn{1}{c}{\textbf{Wan2}} & \multicolumn{1}{c}{\textbf{Kling2}} & \multicolumn{1}{c}{\textbf{Sora}} & \textbf{Avg} \\ \midrule
\multicolumn{1}{c|}{\multirow{5}{*}{\textbf{Base}}} & \multicolumn{1}{l|}{Acc} & 0.946 & 0.931 & 0.967 & 0.960 & 0.976 & 0.964 & 0.944 & 0.947 & 0.971 & \multicolumn{1}{l|}{0.935} & 0.954 \\
\multicolumn{1}{c|}{} & \multicolumn{1}{l|}{Prec} & 0.956 & 0.955 & 0.959 & 0.959 & 0.961 & 0.948 & 0.949 & 0.958 & 0.967 & \multicolumn{1}{l|}{0.952} & 0.956 \\
\multicolumn{1}{c|}{} & \multicolumn{1}{l|}{Recall} & 0.937 & 0.907 & 0.978 & 0.961 & 0.994 & 0.983 & 0.941 & 0.938 & 0.976 & \multicolumn{1}{l|}{0.919} & 0.953 \\
\multicolumn{1}{c|}{} & \multicolumn{1}{l|}{F1} & 0.946 & 0.931 & 0.968 & 0.960 & 0.977 & 0.965 & 0.945 & 0.948 & 0.972 & \multicolumn{1}{l|}{0.935} & 0.955 \\
\multicolumn{1}{c|}{} & \multicolumn{1}{l|}{AUC} & 0.989 & 0.983 & 0.995 & 0.992 & 0.999 & 0.994 & 0.989 & 0.983 & 0.994 & \multicolumn{1}{l|}{0.983} & 0.990 \\ \midrule
\multicolumn{1}{c|}{\multirow{5}{*}{\textbf{Ablation 1}}} & \multicolumn{1}{l|}{Acc} & 0.945 & 0.932 & 0.971 & 0.963 & 0.978 & 0.967 & 0.942 & 0.948 & 0.973 & \multicolumn{1}{l|}{0.936} & 0.955 \\
\multicolumn{1}{c|}{} & \multicolumn{1}{l|}{Prec} & 0.957 & 0.955 & 0.962 & 0.966 & 0.962 & 0.953 & 0.947 & 0.965 & 0.969 & \multicolumn{1}{l|}{0.946} & 0.958 \\
\bottomrule
\end{tabular}%
}
\end{table}

\begin{table}[]
\centering
\caption{Ablation of DINO-v2 test set}
\label{tab:dino2_test}
\resizebox{\textwidth}{!}{%
\begin{tabular}{@{}cllllllllllllllll|l@{}}
\toprule
\multicolumn{1}{l}{\textbf{Model}} & \textbf{Metric} & \multicolumn{1}{c}{\textbf{Unk}} & \multicolumn{1}{c}{\textbf{RM2}} & \multicolumn{1}{c}{\textbf{Kling1}} & \multicolumn{1}{c}{\textbf{Hailuo}} & \multicolumn{1}{c}{\textbf{SD}} & \multicolumn{1}{c}{\textbf{Mochi}} & \multicolumn{1}{c}{\textbf{JM}} & \multicolumn{1}{c}{\textbf{Gen3}} & \multicolumn{1}{c}{\textbf{Luma}} & \multicolumn{1}{c}{\textbf{Vidu}} & \textbf{PRM} & \textbf{SKR} & \textbf{PV} & \textbf{Pika2} & \textbf{Gen4} & \textbf{Avg} \\ \midrule
\multicolumn{1}{c|}{\multirow{5}{*}{\textbf{Base}}} & \multicolumn{1}{l|}{Acc} & 0.920 & 0.973 & 0.828 & 0.963 & 0.958 & \textbf{0.742} & 0.963 & 0.943 & 0.949 & 0.974 & 0.944 & 0.667 & 0.895 & \textbf{0.775} & \textbf{0.621} & 0.874 \\
\multicolumn{1}{c|}{} & \multicolumn{1}{l|}{Prec} & 0.954 & 0.962 & 0.955 & 0.970 & 0.966 & 0.921 & 0.966 & 0.955 & 0.946 & 0.987 & 0.956 & 0.921 & 0.966 & 0.941 & 0.870 & 0.949 \\
\multicolumn{1}{c|}{} & \multicolumn{1}{l|}{Recall} & 0.886 & 0.987 & 0.695 & 0.958 & 0.952 & 0.541 & 0.962 & 0.932 & 0.954 & 0.961 & 0.932 & 0.381 & 0.823 & 0.597 & 0.305 & 0.791 \\
\multicolumn{1}{c|}{} & \multicolumn{1}{l|}{F1} & 0.919 & 0.974 & 0.805 & 0.964 & 0.959 & 0.682 & 0.964 & 0.943 & 0.950 & 0.974 & 0.944 & 0.539 & 0.889 & 0.730 & 0.452 & 0.846 \\
\multicolumn{1}{c|}{} & \multicolumn{1}{l|}{AUC} & 0.975 & 0.997 & 0.935 & 0.992 & 0.990 & 0.882 & 0.992 & 0.984 & 0.991 & 0.997 & 0.984 & 0.886 & 0.972 & 0.923 & 0.862 & 0.957 \\ \midrule
\multicolumn{1}{c|}{\multirow{5}{*}{\textbf{Ablation 1}}} & \multicolumn{1}{l|}{Acc} & 0.923 & 0.975 & 0.845 & 0.963 & 0.948 & \textbf{0.735} & 0.964 & 0.941 & 0.951 & 0.963 & 0.932 & \textbf{0.667} & 0.893 & \textbf{0.791} & \textbf{0.638} & 0.875 \\
\multicolumn{1}{c|}{} & \multicolumn{1}{l|}{Prec} & 0.961 & 0.964 & 0.967 & 0.973 & 0.962 & 0.924 & 0.964 & 0.959 & 0.950 & 0.979 & 0.949 & 0.926 & 0.962 & 0.951 & 0.909 & 0.953 \\
\bottomrule
\end{tabular}%
}
\end{table}

\end{document}